\documentclass[journal]{IEEEtran}

\usepackage{amsmath}
\usepackage{graphicx}
\usepackage{times}
\usepackage{caption2}

\usepackage{amsmath}

\usepackage{amsfonts}
\usepackage{amssymb}
\usepackage[mathscr]{euscript}
\usepackage{comment}
\usepackage{bm}
\usepackage{subfigure}
\usepackage{algorithm}
\usepackage{algorithmic}
\usepackage{multirow}
\usepackage{wrapfig}
\usepackage{booktabs}
\usepackage{threeparttable}
\usepackage[english]{babel}
\usepackage{microtype}
\usepackage{url}

\usepackage[bookmarks,backref=true,linkcolor=black]{hyperref}
\hypersetup{
  pdfauthor = {},
  pdftitle = {},
  pdfsubject = {},
  pdfkeywords = {},
  colorlinks=true,
  linkcolor= black,
  citecolor= black,
  pageanchor=true,
  urlcolor = black,
  plainpages = false,
  linktocpage
}
\usepackage{color}

\newcommand{\bfA}{\mathbf{A}}
\newcommand{\bfb}{\mathbf{b}}

\newcommand{\bfd}{\mathbf{d}}

\newcommand{\bfn}{\mathbf{n}}
\newcommand{\bfp}{\mathbf{p}}

\newcommand{\bfq}{\mathbf{q}}
\newcommand{\bfr}{\mathbf{r}}

\newcommand{\bft}{\mathbf{t}}

\newcommand{\bfw}{\mathbf{w}}

\newcommand{\bfz}{\mathbf{z}}
\newcommand{\bfalpha}{\bm{\alpha}}
\newcommand{\bfsigma}{\bm{\sigma}}

\newcommand{\txid}{\textrm{id}}
\newcommand{\txexp}{\textrm{exp}}
\newcommand{\txalb}{\textrm{alb}}

\newcommand*\Laplace{\mathop{}\!\mathbin\bigtriangleup}

\newcommand{\para}[1]{\textbf{{#1}.}}

\definecolor{turquoise}{cmyk}{0.65,0,0.1,0.1}
\definecolor{purple}{rgb}{0.65,0,0.65}
\definecolor{dark_green}{rgb}{0, 0.5, 0}
\definecolor{orange}{rgb}{0.8, 0.2, 0.2}

\newcommand{\juyong}[1]{#1}

\newcommand{\edit}[1]{{\color{black}#1}}
\newcommand{\editnew}[1]{{\color{black}#1}}

\begin{document}

\title{\edit{CNN-based Real-time Dense Face Reconstruction with Inverse-rendered Photo-realistic Face Images}}

\author{Yudong~Guo, Juyong~Zhang$^\dagger$, Jianfei~Cai, Boyi~Jiang and~Jianmin~Zheng
\thanks{Yudong~Guo, Juyong~Zhang and Boyi Jiang are with School of Mathematical Sciences, University of Science and Technology of China.}
\thanks{Jianfei~Cai and Jianmin~Zheng are with School of Computer Science and Engineering, Nanyang Technological University.}
\thanks{$^\dagger$Corresponding author. Email: \href{mailto:juyong@ustc.edu.cn}{\texttt{juyong@ustc.edu.cn}}.}
}

\maketitle

\begin{abstract}

With the powerfulness of convolution neural networks (CNN), CNN based face reconstruction has recently shown promising performance in reconstructing detailed face shape from 2D face images. The success of CNN-based methods relies on a large number of labeled data. The state-of-the-art synthesizes such data using a coarse morphable face model, which however has difficulty to generate detailed photo-realistic images of faces (with wrinkles). This paper presents a novel face data generation method. Specifically, we render a large number of photo-realistic face images with different attributes based on inverse rendering. Furthermore, we construct a fine-detailed face image dataset by transferring different scales of details from one image to another. We also construct a large number of video-type adjacent frame pairs by simulating the distribution of real video data~\footnote{All these coarse-scale and fine-scale photo-realistic face image datasets can be downloaded from \url{https://github.com/Juyong/3DFace}.}. With these nicely constructed datasets, we propose a coarse-to-fine learning framework consisting of three convolutional networks. The networks are trained for real-time detailed 3D face reconstruction from monocular video as well as from a single image. Extensive experimental results demonstrate that our framework can produce high-quality reconstruction but with much less computation time compared to the state-of-the-art. Moreover, our method is robust to pose, expression and lighting due to the diversity of data.
\end{abstract}

\begin{IEEEkeywords}
3D face reconstruction, face tracking, face performance capturing, 3D face dataset, image synthesis, deep learning
\end{IEEEkeywords}

\section{Introduction} \label{se:intro}
This paper considers the problem of dense 3D face reconstruction from monocular video as well as from a single face image. Single-image based 3D face reconstruction can be considered as a special case of video based reconstruction. It also plays an essential role.
Actually image-based 3D face reconstruction itself is a fundamental problem in computer vision and graphics, and has many applications such as face recognition~\cite{blanz2003face, tran2017regressing} and face animation \cite{ichim2015dynamic, thies2016face2face}.  Video-based dense face reconstruction and tracking or facial performance capturing has a long history~\cite{williams1990performance} also with many applications such as facial expression transfer ~\cite{thies2015realtime, thies2016face2face} and face replacement~\cite{kemelmacher2010being,dale2011video,garrido2014automatic}. Traditional facial performance capture methods usually require complex hardware and significant user intervention~\cite{williams1990performance,huang2011leveraging} to achieve a sufficient reality and therefore are not suitable for consumer-level applications. Commodity RGB-D camera based methods~\cite{weise2011realtime,li2013realtime,bouaziz2013online,thies2015realtime} have demonstrated real-time reconstruction and animation results. However, RGB-D devices, such as Microsoft's Kinect, are still not that common and not of high resolution,  compared to RGB devices.

Recently, several approaches have been proposed for RGB video based facial performance captureing~\cite{cao2014displaced,cao2015real,thies2016face2face,garrido2016reconstruction,saito2016real,huber2016multiresolution}.
Compared to image-based 3D face reconstruction that is considered as an ill-pose and challenging task due to the ambiguities caused by insufficient information conveyed in 2D images, video-based 3D reconstruction and tracking is even more challenging especially when the reconstruction is required to be real-time, fine-detailed and robust to pose, facial expression, lighting, etc. These proposed approaches only partially comply with the requirements. For example, \cite{cao2014displaced} and \cite{cao2015real} learn facial geometry while not recovering facial appearance property, such as albedo. ~\cite{garrido2016reconstruction} can reconstruct personalized face rig of high-quality, but their optimization-based method is time-consuming and needs about 3 minutes per frame. ~\cite{thies2016face2face} achieves real-time face reconstruction and facial reenactment through data-parallel optimization strategy, but their method cannot recover fine-scale details such as wrinkles and also requires facial landmark inputs.

In this paper, we present a solution to tackle all these problems by utilizing the powerfulness of convolutional neural networks (CNN). CNN based approaches have been proposed for face reconstruction from a single image~\cite{richardson20163d,richardson2016learning,tran2017regressing,tewari17MoFA,jackson2017large}, but CNN is rarely explored for video-based dense face reconstruction and tracking, especially for real-time reconstruction. Inspired by the state-of-the-art single-image based face reconstruction method~\cite{richardson2016learning}, which employs two cascaded CNNs (coarse-layer CNN and fine-layer CNN) to reconstruct a detailed 3D facial surface from a single image, we develop a dense face reconstruction and tracking framework. The framework includes a new network architecture called 3DFaceNet for online real-time dense face reconstruction from monocular video (supporting a single-image input as well), and optimization-based inverse rendering for offline generating large-scale training datasets.

In particular, our proposed 3DFaceNet consists of three convolutional networks: a coarse-scale single-image network (named Single-image CoarseNet for the first frame or the single image case), a coarse-scale tracking network (Tracking CoarseNet) and a fine-scale network (FineNet). For single-image based reconstruction, compared with~\cite{richardson2016learning}, the key
uniqueness of our framework lies in the photo-realistic datasets
we generate for training CoarseNet and FineNet.

It is known that one major challenge for CNN-based methods lies in the difficulty to obtain a large number of labelled training data. For our case, there is no publicly available dataset that can provide large-scale face images with their corresponding high-quality 3D face models. For training CoarseNet, \cite{richardson20163d} and \cite{richardson2016learning} resolve the training data problem by directly synthesizing face images with randomized parametric face model parameters. Nevertheless, due to the low dimensionality of the parametric face model, albedo and random background synthesized, the rendered images in~\cite{richardson20163d, richardson2016learning} are not photo-realistic. In contrast, we propose to create realistic face images by starting from real photographs and manipulating them after an inverse rendering procedure. For training FineNet, because of no dataset with detailed face geometry, \cite{richardson2016learning} uses an unsupervised training by adopting the shading energy as the loss function. However, to make back-propagation trackable, \cite{richardson2016learning} employs the first-order spherical harmonics to model the lighting, which makes the final detailed reconstruction not so accurate. On the contrary, we propose a novel approach to transfer different scales of details from one image to another. With the constructed fine-detailed face image dataset, we can train FineNet in a fully supervised manner, instead of the unsupervised way in~\cite{richardson2016learning}, and thus can produce more accurate reconstruction results. Moreover, for training our coarse-scale tracking network for the video input case, we consider the coherence between adjacent frames and simulate adjacent frames according to the statistics learned from real facial videos for training data generation.

\para{Contributions} In summary, the main contributions of this paper lie in the following five aspects:
\begin{itemize}
\item the optimization-based face inverse rendering  that recovers accurate
geometry, albedo, lighting from a single image, with which we can generate a large number of photo-realistic
face images with different attributes to train our networks.
\item a large photo-realistic face image dataset with the labels of the parametric face model parameters and the pose parameters,
which are generated based on our proposed inverse rendering. This dataset facilitates the training of
our Single-image CoarseNet and makes our method robust to expressions and poses.
\item a large photo-realistic fine-scale face image dataset with detailed geometry labels, which are generated by
our proposed face detail transfer approach. This fine-scale dataset facilitates the training of our FineNet.
\item a large dataset for training Tracking CoarseNet, where we extend the Single-image CoarseNet training data by simulating their previous frames according to the statistics learned from real facial videos.
\item the proposed 3DFaceNet that is trained with our built large-scale diverse synthetic data and is thus able to reconstruct the fine-scale geometry,
albedo and lighting well in real time from monocular
RGB video as well a single image. Our system is robust to large poses,
extreme expressions and fast moving faces.
\end{itemize}
To the best of our knowledge, the proposed framework is the first work that achieves real-time dense 3D face reconstruction and tracking from monocular video. It might open up a new venue of research in the field of 3D assisted face video analysis. Moreover, the optimization-based face inverse rendering approach provides a novel, efficient way to generate various large-scale synthetic dataset by appropriate adaptation. Our elaborately-generated datasets will also benefit the face analysis related research that usually requires large amounts of training data.

\section{Related Work}
3D face reconstruction and facial performance capturing have been studied extensively in computer vision and computer graphics communities. For conciseness, we only review the most relevant works here.

\para{Low-dimensional Face Models} Model-based approaches for face shape reconstruction have grown in popularity over the last decade. Blanz and Vetter \cite{blanz1999face} proposed to represent a textured 3D face with principal components analysis (PCA), which provides an effective low-dimensional representation in terms of latent variables and corresponding basis vectors~\cite{stewart1993early}. The model has been widely used in various computer vision tasks, such as face recognition~\cite{blanz2003face, tran2017regressing}, face alignment~\cite{zhu2016face, joint-face-alignment, large-pose-face-alignment}, and face reenactment~\cite{thies2016face2face}. Although such a model is able to capture the global structure of a 3D face from a single image \cite{blanz1999face} or multiple images \cite{amberg2007reconstructing}, the facial details like wrinkles and folds are not possible to be captured. In addition, the reconstructed face models rely heavily on training samples. For example, a face shape is difficult to be reconstructed if it is far away from the span of the training samples. Thus, similar to~\cite{richardson2016learning}, we only use the low-dimensional model in our coarse layer to reconstruct a rough geometry and we refine the geometry in our fine layer.

\para{Shape-from-shading (SFS)}
SFS~\cite{prados2006shape} makes use of the rendering principle to recover the underlying shape from shading observations. The performance of SFS largely depends on constraints or priors. For 3D face reconstruction, in order to achieve plausible results, the prior knowledge about the geometry must be applied. For instance, in order to reduce the ambiguity and the complexity of SFS, the symmetry of the human face has often been employed~\cite{shimshoni2000shape, zhao2000illumination, zhao2001symmetric}. Kemelmacher et al.~\cite{kemelmacher20113d} used a reference model prior to align with the face image and then applied SFS to refine the reference model to better match the image. Despite the improved performance of this technique, its capability to capture global face structure is limited.

\para{Inverse Rendering} The generation of a face image depends on several factors: face geometry, albedo, lighting, pose and camera parameters. Face inverse rendering refers to the process of estimating all these factors from a real face image, which can then be manipulated to render new images. Inverse rendering is similar to SFS with the difference that inverse rendering aims to estimate all the rendering parameters while SFS mainly cares about reconstructing the geometry. Aldrian et al.~\cite{aldrian2013inverse} did face inverse rendering with a parametric face model using a multilinear approach, where the face geometry and the albedo are encoded on parametric face model. In~\cite{aldrian2013inverse}, the geometry is first estimated based on the detected landmarks, and then the albedo and the lighting are iteratively estimated by solving the rendering equation. However, since the landmark constraint is a sparse constraint, the reconstructed geometry may not fit the face image well. \cite{garrido2016reconstruction} fits a 3D face in a multi-layer approach and extracts a high-fidelity parameterized 3D rig that contains a generative wrinkle formation model capturing the person-specific idiosyncrasies. \edit{~\cite{bas2016fitting} presents an algorithm for fully automatically fitting a 3D Morphable Model to a single image using landmarks and edge features. ~\cite{schonborn2017markov} introduces a framework to fit a parametric face model  with Bayesian inference. ~\cite{egger2016occlusion} and \cite{Egger2018} estimate an occlusion map and fit a statistical model to a face image with an EM-like probabilistic estimation process.} \cite{JiangZDLL17} adopts the similar approach to recover the 3D face model with geometry details. While these methods provide impressive results, they are usually time-consuming due to complex optimization.

\para{Face Capture from RGB Videos} Recently, a variety of methods have been proposed to do 3D face reconstruction with monocular RGB video. Most of them use a 3D Morphable Model~\cite{thies2016face2face,garrido2016reconstruction,huber2016multiresolution} or a multi-linear face model~\cite{cao20133d,cao2014displaced,shi2014automatic,cao2015real,saito2016real} as a prior. ~\cite{garg2013dense} reconstructs the dense 3D face from a monocular video sequence by a variational approach, which is formulated as estimating dense low-rank smooth 3D shapes for each frame of the video sequence. ~\cite{GVWT13} adapts a generic template to a static 3D scan of an actor's face, then fits the blendshape model to monocular video off-line, and finally extracts surface detail by shading-based shape refinement under general lighting. ~\cite{shi2014automatic} uses a similar tracking approach and achieves impressive results based on global energy optimization of a set of selected keyframes. ~\cite{garrido2016reconstruction} fits a 3D face in a multi-layer approach and extracts a high-fidelity parameterized 3D rig that contains a generative wrinkle formation model capturing the person-specific idiosyncrasies. Although all these methods provide impressive results, they are time-consuming and are not suitable for real-time face video reconstruction and editing. ~\cite{cao20133d,cao2014displaced} adopt a learning-based regression model to fit a generic identity and expression model to a RGB face video in real-time and~\cite{cao2015real} extends this approach by also regressing fine-scale face wrinkles. \edit{~\cite{saito2016real} presents a method for unconstrained real-time 3D facial performance capture through explicit semantic segmentation in the RGB input. ~\cite{huber2016multiresolution} tracks face by fitting 3D Morphable Model to the detected landmarks.} Although they are able to reconstruct and track 3D face in real-time, they do not estimate facial appearance. Recently,~\cite{thies2016face2face} presented an approach for real-time face tracking and facial reenactment, but the method is not able to recover fine-scale details and requires external landmark inputs. In contrast, our method is the first work that can do real-time reconstruction of face geometry at fine details as well as real-time recovery of albedo, lighting and pose parameters.

\para{Learning-based Single-image 3D Face Reconstruction} With the powerfulness of convolution neural networks, deep learning based methods have been proposed to do 3D face reconstruction from one single image. ~\cite{large-pose-face-alignment,zhu2016face,KimZTTRT17} use 3D Morphable Model (3DMM) ~\cite{blanz1999face} to represent 3D faces and use CNN to learn the 3DMM and pose parameters. ~\cite{richardson20163d} follows the method and uses synthetic face images generated by rendering textured 3D faces encoded on 3DMM with random lighting and pose for training data. However, the reconstruction results of these methods do not contain geometry details. Besides learning the 3DMM and pose parameters,~\cite{richardson2016learning} extends these methods by also learning detailed geometry in an unsupervised manner. ~\cite{tran2017regressing} proposes to regress robust and discriminative 3DMM with a very deep neural network and uses it for face recognition. ~\cite{tewari17MoFA} proposes to use an analysis-by-synthesis energy function as the loss function during network training~\cite{blanz1999face,thies2016face2face}. \edit{\cite{jackson2017large} proposes to directly regress volumes with CNN for a single face image.} Although these methods utilize the powerfulness of CNNs, they all concentrate on images and do not account for videos. In comparison, we focus on monocular face video input and reconstruct face video in real-time by using CNNs.

\section{Face Rendering Process} \label{se:rendering}
This section describes some background information, particularly on the face representations and the face rendering process considered in our work.
The rendering process of a face image depends on several factors: face geometry, albedo, lighting, pose and camera parameters. We encode 3D face geometry into two layers: a coarse-scale shape and fine-scale details. While the coarse-scale shape and albedo are represented by a parametric textured 3D face model, the fine-scale details are represented by a pixel depth displacement map. The face shape is represented via a mesh of $n$ vertices with fixed connectivity as a vector $\bfp = [\bfp^T_1, \bfp^T_2, \ldots, \bfp^T_{n}]^T\in \mathcal{R}^{3n}$, where $\bfp_i$ denotes the position of vertex $v_i$ ($i = 1, 2, \ldots, n$).\\

\para{Parametric face model}
We use 3D Morphable Model (3DMM)~\cite{blanz1999face} as the parametric face model to encode 3D face geometry and albedo on a lower-dimensional subspace, and extend the shape model to also cover facial expressions by adding delta blendshapes. Specifically, the parametric face model describes 3D face geometry $\bfp$ and albedo $\bfb$ with PCA (principle component analysis):
\begin{equation}
\bfp = \bar{\bfp} + \bfA_{\txid}\bfalpha_{\txid} + \bfA_{\txexp}\bfalpha_{\txexp} ,
\label{eq:3dmm_geo}
\end{equation}
\begin{equation}
\bfb = \bar{\bfb} + \bfA_{\txalb}\bfalpha_{\txalb},
\label{eq:3dmm_tex}
\end{equation}
where $\bar{\bfp}$ and $\bar{\bfb}$ denote respectively the shape and the albedo of
the average 3D face, $\bfA_{\txid}$ and $\bfA_{\txalb}$ are the principle axes extracted from a set of textured 3D meshes with a neutral expression, $\bfA_{\txexp}$ represents the principle axes trained on the offsets between the expression meshes and the neutral meshes of individual persons, and $\bfalpha_{\txid}$, $\bfalpha_{\txexp}$ and $\bfalpha_{\txalb}$ are the corresponding coefficient vectors that characterize a specific 3D face model. For diversity and mutual complement, we use the Basel Face Model (BFM)~\cite{paysan20093d} for $\bfA_{\txid}$ and $\bfA_{\txalb}$ and FaceWarehouse~\cite{cao2014facewarehouse} for $\bfA_{\txexp}$.\\

\para{Fine-scale details}
As 3DMM is a low-dimensional model, some face details such as wrinkles and dimples cannot be expressed by 3DMM. Thus, we encode the geometry details in a displacement along the depth direction for each pixel.\\

\para{Rendering process}
For camera parametrization, following~\cite{richardson2016learning}, we use the weak perspective model to project the 3D face onto the image plane:
\begin{equation}
\bfq_i
=
s
\left(
\begin{array}{ccc}
1 & 0 & 0\\
0 & 1 & 0\\
\end{array}
\right)
 R\bfp_i
+
\bft,
\label{eq:project}
\end{equation}
where $\bfp_i$ and $\bfq_i$ are the locations of vertex $v_i$ in the world coordinate system and in the image plane, respectively, $s$ is the scale factor, $R$ is the rotation matrix constructed from Euler angles $pitch, yaw, roll$ and $\bft = (t_x, t_y)^T$ is the translation vector.

To model the scene lighting, we assume the face to be a Lambertian surface. The global illumination is approximated using the spherical harmonics (SH) basis functions~\cite{sphereharmonics}. Then, the irradiance of a vertex $v_i$ with surface normal $\bfn_i$ and scalar albedo $b_i$ is expressed as~\cite{ramamoorthi2001efficient}:
\begin{equation}
\mathcal{L}(\bfn_i, b_i \ | \  \gamma) = b_i\cdot\sum\limits_{k = 1}^{B^{2}}\gamma_{k}\phi_{k}(\bfn_i),
\label{eq:irradiance}
\end{equation}
where $\bm{\phi}(\bfn_i) = [\phi_{1}(\bfn_i),\ldots, \phi_{B^2}(\bfn_i)]^{T}$ is the SH basis functions computed with normal $\bfn_i$, and $\bm{\gamma} = [\gamma_{1},\ldots,\gamma_{B^{2}}]^{T}$ is the SH coefficients. We use the first $B = 3$ bands of SHs for the illumination model. Thus, the rendering process depends on the parameter set $\chi = \{\bfalpha_{\txid}, \bfalpha_{\txexp}, \bfalpha_{\txalb}, s, pitch, yaw, roll, \bft, \bfr\}$, where $\bfr = (\bm{\gamma}_{r}^{T}, \bm{\gamma}_{g}^{T}, \bm{\gamma}_{b}^{T})^{T}$ denotes RGB channels' SH illumination coefficients.

Given the parametric face model and the parameter set $\chi$, a face image can be rendered as follows. First, a textured 3D mesh is constructed using Eq.~\eqref{eq:3dmm_geo} and Eq.~\eqref{eq:3dmm_tex}. Then we do a rasterization via Eq.~\eqref{eq:project}. Particularly, in the rasterization, for every pixel in the face region of the 2D image, we obtain the underlying triangle index on the 3D mesh and its barycentric coordinates. In this way, for every pixel in the face region, we obtain its normal by using the underlying triangle's normal, and its albedo value by barycentrically interpolating the albedos of the vertices of the underlying triangle. Finally, with the normal, the albedo and the lighting, the color of a pixel can be rendered using Eq.~\eqref{eq:irradiance}.

\section{Overview of Proposed Learning-based Dense Face Reconstruction}

\begin{figure}[htbp]
      \centering
	\includegraphics[width=1\columnwidth]{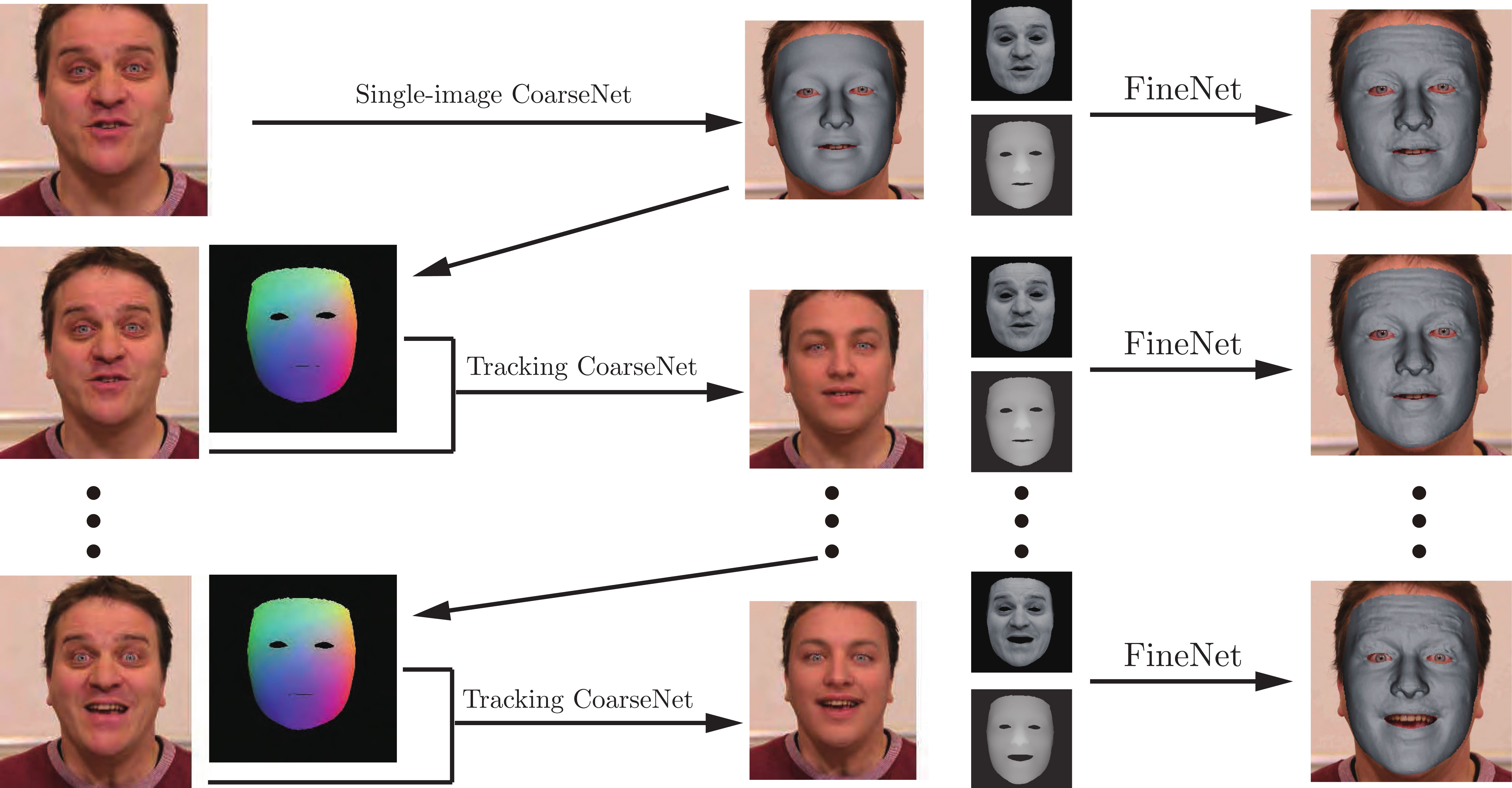}
      \caption{The pipeline of our proposed learning based dense 3D face reconstruction and tracking framework. The first frame of the input video is initially reconstructed by a single-image CoarseNet for coarse face geometry reconstruction, followed by using FineNet for detailed face geometry recovery. Each of the subsequent frames is processed by a tracking CoarseNet followed by FineNet.}
      \label{fig:LearningPipeline}
\end{figure}

To achieve real-time face video reconstruction and tracking, we need real-time face inverse rendering. However, reconstructing detailed 3D face using traditional optimization-based methods~\cite{garrido2016reconstruction} is far from real-time. To address this problem, we develop a novel CNN based framework to achieve real-time detailed face inverse rendering. Specifically, we use two CNNs for each frame, namely CoarseNet and FineNet. The first one estimates coarse-scale geometry, albedo, lighting and pose parameters altogether, and the second one reconstructs the fine-scale geometry encoded on pixel level.

Fig.~\ref{fig:LearningPipeline} shows the entire system pipeline. It can be seen that there are two types of CoarseNet: Single-image CoarseNet and Tracking CoarseNet. Tracking CoarseNet makes use of the predicted parameters of the previous frame, while Single-image CoarseNet is for the first frame case where there is no previous frame available. Such Single-image CoarseNet could be applied to other key frames as well to avoid any potential drifting problem if needed. The combination of all the networks including Single-image CoarseNet, Tracking CoarseNet and FineNet, makes up a complete framework for real-time dense 3D face reconstruction from monocular video. Note that the entire framework can be easily degenerated to the solution for dense 3D face reconstruction from a single image by combining only Single-image CoarseNet with FineNet.

We would like to point out that although we advocate the CNN based solution, it still needs to work together with optimization based inverse rendering methods. This is because CNN requires large amount of data with labels, which is usually not available, and optimization based inverse rendering methods are a natural solution for generating labels (optimal parameters) and synthesizing new images offline. Thus, our proposed dense face reconstruction and tracking framework includes both optimization based inverse rendering and the two-stage CNN based solution, where the former is for offline training data generation and the latter is for real-time online operations. In the subsequent sections, we first introduce our optimization based inverse face rendering, which will be used to construct training data for CoarseNet and FineNet; and then we present our three convolutional networks.

\section{Optimization Based Face Inverse Rendering}
\label{se:inverse_rendering}
\begin{figure*}
  \centering
\includegraphics[width=1\textwidth]{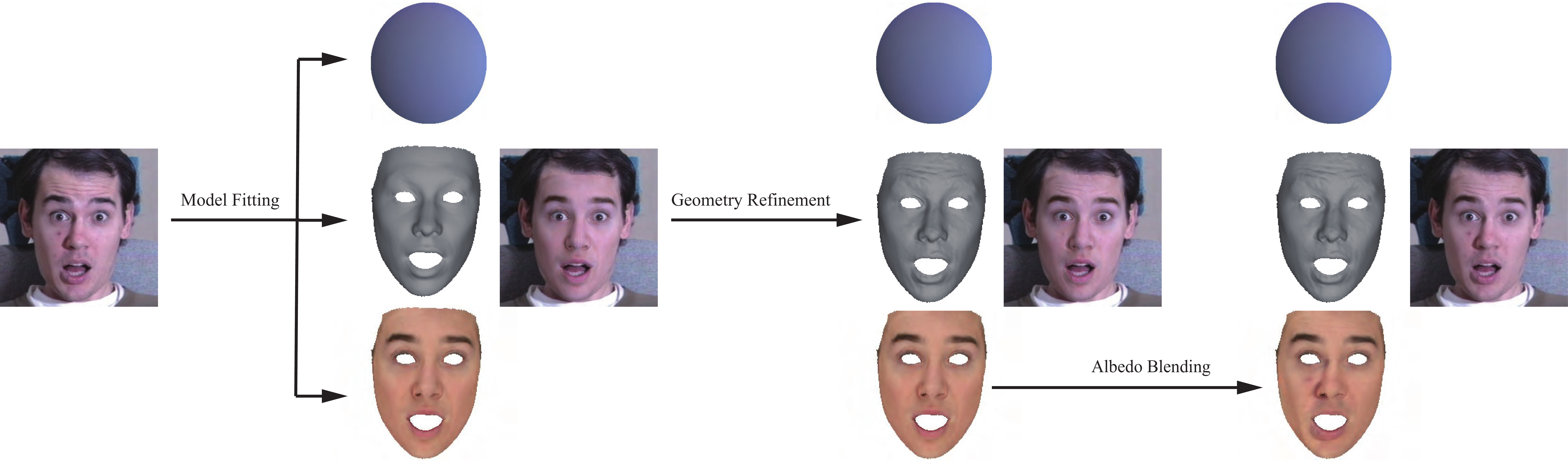}
  \caption{The pipeline of our proposed inverse rendering method. Given an input face image (left), our inverse rendering consists of three stages: Model fitting (second column), geometry refinement (third column) and albedo blending (last column). At each stage, the top to bottom rows are the corresponding recovered lighting, geometry and albedo, and the rendered face image is shown on the right. The arrows indicate which component is updated.}\label{fig:inverserender}
\end{figure*}

Inverse rendering is an inverse process of image generation. That is, given a face image, we want to estimate a 3D face with albedo, lighting condition, pose and projection parameters simultaneously. Since directly estimating these unknowns with only one input image is an ill-posed problem, we use the parametric face model as a prior. Fig.~\ref{fig:inverserender} illustrates our developed inverse rendering, which consists of three stages: parametric face model fitting, geometry refinement and albedo blending. The first stage is to recover the lighting, a coarse geometry and the albedo based on the parametric face model. The second stage is to further recover the geometry details. The third stage is to blend the albedo so as to make the rendered image closer to the input image. Via the developed inverse rendering, we are able to extract different rendering components of real face images, and then by varying these different components we can create large-scale photo-realistic face images to facilitate the subsequent CNN based training.

\subsection{Stage 1 - Model Fitting} \label{se:model_fitting}
The purpose of model fitting is to estimate the coarse face geometry, albedo, lighting, pose and projection parameters from a face image $I_{\textrm{in}}$. That is to estimate $\chi = \{\bfalpha_{\txid}, \bfalpha_{\txexp}, \bfalpha_{\txalb}, s, pitch, yaw, roll, \bft, \bfr\}$. For convenience, we group these parameters into the following sets $\chi_{g} = \{\bfalpha_{\txid}, \bfalpha_{\txexp}\}$, $\chi_{p} = \{pitch, yaw, roll\}$, $\chi_{t} = \{s,  \bft\}$ and $\chi_{l} = \{\bfalpha_{\txalb}, \bfr\}$. The fitting process is based on the analysis-by-synthesis strategy~\cite{blanz1999face,thies2016face2face}, and we seek a solution that by minimizes the difference between the input face image and the rendered image with $\chi$. Specifically, we minimize the following objective function:
\begin{equation}
E(\chi) = E_{\textrm{con}} + w_{l}E_{\textrm{lan}} + w_{r}E_{\textrm{reg}},
\label{eq:3dmm_object}
\end{equation}
where $E_{\textrm{con}}$ is a photo-consistency term, $E_{\textrm{lan}}$ is a landmark term and $E_{\textrm{reg}}$ is a regularization term, and $w_{l}$ and $w_{r}$ are tradeoff parameters. The photo-consistency term, aiming to minimize the difference between the input face image and the rendered image, is defined as
\begin{equation}
E_{\textrm{con}}(\chi) = \frac{1}{|\mathcal{F}|}\|I_{\textrm{ren}} - I_{\textrm{in}}\|^2,
\end{equation}
where $I_{\textrm{ren}}$ is the rendered image, $I_{\textrm{in}}$ is the input image, and $\mathcal{F}$ is the set of all pixels in the face region. The landmark term aims to make the projected vertices close to the corresponding landmarks in the image plane:
\begin{equation}
E_{\textrm{lan}}(\chi) = \frac{1}{|\mathcal{L}|}\sum_{i \in \mathcal{L}}^{}\|\bfq_{i} - (\Pi R\bfp_{i} + \bft)\|^2,
\label{eq:land}
\end{equation}
where $\mathcal{L}$ is the set of landmarks, $\bfq_{i}$ is a landmark position in the image plane, $\bfp_i$ is the corresponding vertex location in the fitted 3D face and $\Pi = s
\left(
\begin{array}{ccc}
1 & 0 & 0\\
0 & 1 & 0\\
\end{array}
\right)$. The regularization term aims to ensure that the fitted parametric face model parameters are plausible:
\begin{equation}
E_{\textrm{reg}}(\chi)=\sum_{i=1}^{100}\left[\left(\frac{\bfalpha_{\txid, i}}{\bfsigma_{\txid,i}}\right)^2+\left(\frac{\bfalpha_{\txalb, i}}{\bfsigma_{\txalb,i}}\right)^2\right] + \sum_{i = 1}^{79}\left(\frac{\bfalpha_{\txexp, i}}{\bfsigma_{\txexp,i}}\right)^2,
\end{equation}
where $\bfsigma$ is the standard deviation of the corresponding principal direction. Here we use 100 principle components for identity $\&$ albedo, and 79 for expression. In our experiments, we set $w_{l}$ to be 10 and $w_{r}$ to be $5 \cdot 10^{-5}$. Eq.~\eqref{eq:3dmm_object} is minimized via Gauss-Newton iteration.

\subsection{Stage 2 - Geometry Refinement} \label{se:geo_refine}
As the parametric face model is a low-dimensional model, some face details such as wrinkles and dimples are not encoded in parametric face model. Thus, the purpose of the second stage is to refine the geometry by adding the geometry details in a displacement along the depth direction for every pixel. In particular, by projecting the fitted 3D face with parameter $\chi$, we can obtain a depth value for every pixel in the face region. Let $\bfz$ be all stacked depth values of pixels, $\bfd$ be all stacked displacements and $\widetilde{\bfz} = \bfz + \bfd$ be all new depth values. Given new depth values $\widetilde{\bfz}$, the normal at pixel $(i, j)$ can be computed using the normal of triangle $(\bfp_{(i, j)}, \bfp_{(i + 1,  j)}, \bfp_{(i,  j + 1)})$, where $\bfp_{(i, j)} = [i, j, \widetilde{\bfz}(i, j)]^{T}$ is the coordinates of pixel $(i, j)$ at the camera system. Inspired by \cite{ichim2015dynamic}, we estimate $\bfd$ using the following objective function:
\begin{equation}
E(\bfd) = E_{\textrm{con}} + \mu_{1}\|\bfd\|_{2}^2 + \mu_{2}\|\Laplace\bfd\|_{1},
\label{eq:displacement}
\end{equation}
where $E_{\textrm{con}}$ is the same as that in Eq.~\eqref{eq:3dmm_object}, $\|\bfd\|_{2}^2$ is to encourage small displacements, the Laplacian of displacements $\Laplace\bfd$ is to make the displacement smooth, and $\mu_{1}$ and $\mu_{2}$ are tradeoff parameters. We use $\ell_{1}$ norm for the smooth term as it allows preserving sharp discontinuities while removing noise. We set $\mu_{1}$ to be $1 \cdot 10^{-3}$ and $\mu_{2}$ to be 0.3 in our experiments. Eq.~\eqref{eq:displacement} is minimized by using an iterative reweighing approach~\cite{chartrand2008iteratively}.

\subsection{Stage 3 - Albedo Blending} \label{se:albedo_blend}
Similar to the geometry, the albedo encoded in the parametric face model (denoted as $\bfb_{c}$) in stage 1 is also smooth because of the low dimension. For photo-realistic rendering, we extract a fine-scale albedo as
\begin{equation}
\bfb_{f} = I_{\textrm{in}}./(\bfr^{T}\bm{\phi}(\bfn)),
\end{equation}
where $./$ represents the elementwise division operation, $I_{\textrm{in}}$ is the color of the input image and $\bfn$ is the normal computed from the refined geometry. However, the fine-scale albedo $\bfb_{f}$ might contain some geometry details due to imperfect geometry refinement. To avoid this, we linearly blend $\bfb_{c}$ and $\bfb_{f}$, i.e. $\beta \bfb_{c} + (1 - \beta) \bfb_{f}$, with different weights $\beta$ at different regions. Particularly, in the regions where geometry details are likely to appear such as forehead and eye corners, we make the blended albedo close to $\bfb_{c}$ by setting $\beta$ to be 0.65, while in the other regions we encourage  the blended albedo close to $\bfb_{f}$ by setting $\beta$ to be 0.35. Around the border of the regions $\beta$ is set continuously from 0.35 to 0.65. Finally, we use this blended albedo as $\bfb$ in Eq.~\eqref{eq:irradiance} for our subsequent data generation process.

\section{Single-image CoarseNet for Coarse Reconstruction from A Single Image}
In this section, we describe how to train a coarse-layer CNN (called Single-image CoarseNet) that can output the parametric face model parameters (corresponding to a coarse shape) and the pose parameters from the input of a single face image or an independent video frame. Although the network structure of Single-image CoarseNet is similar to that of~\cite{zhu2016face,richardson2016learning}, we use our uniquely constructed training data and loss function, which are elaborated below.

\subsection{Constructing Single-image CoarseNet Training Data} \label{se:coarse_data}
To train Single-image CoarseNet, we need a large-scale dataset of face images with ground-truth 3DMM parameters and pose parameters. Recently, \cite{zhu2016face} proposed to synthesize a large number of face images by varying the 3DMM parameters fitted from a small number of real face images. \cite{zhu2016face} focuses on the face alignment problem. The color of the synthesized face images are directly copied from the source images without considering the underlying rendering process, which makes the synthesized images not photo-realistic and thus unsuitable for high-quality 3D face reconstruction. Later, \cite{richardson2016learning} follows the idea of using synthetic data for learning detailed 3D face reconstruction and directly renders a large number of face images by varying the existing 3DMM parameters with random texture, lighting, and reflectance. However, since 3DMM is a low-dimensional model and the albedo is also of low frequency, the synthetic images in~\cite{richardson2016learning} are not photo-realistic as well, not to mention the random background used in the rendered images. In addition, the synthetic images in~\cite{richardson2016learning} are not available to the public.

\begin{figure}
  \centering
  \includegraphics[width=0.5\textwidth]{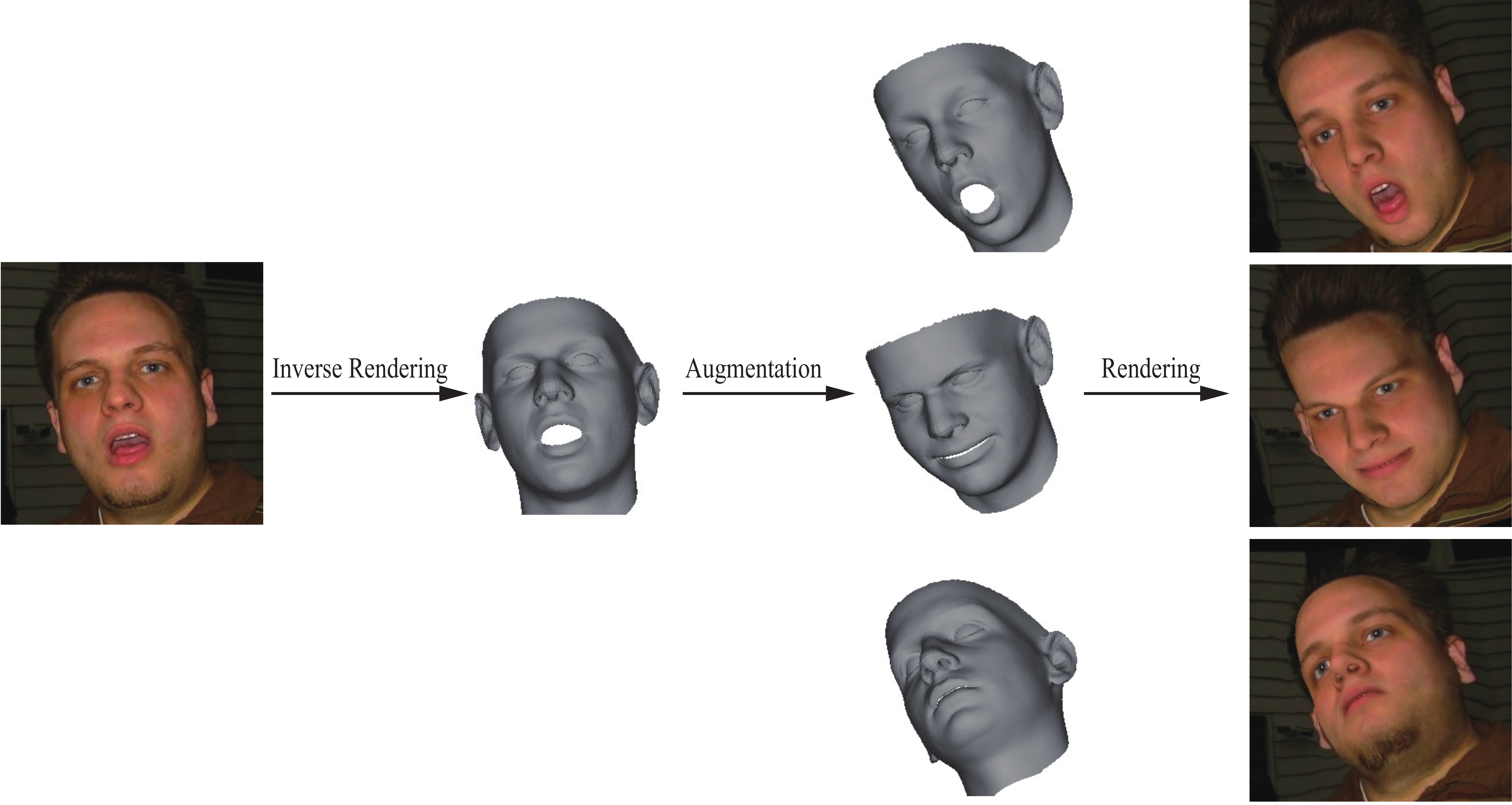}
  \caption{Training data synthesis for Single-image CoarseNet. Given a real face image, we first do the inverse rendering to estimate lighting, albedo and geometry. Then, by changing the expression parameter $\alpha_{exp}$ and the pose parameters $pitch, yaw, roll$, the face geometry is augmented. In the final, a set of new face images is obtained by rendering the newly changed face geometry.}\label{fig:coarsedata}
\end{figure}

Therefore, in this paper, we propose to use our developed inverse rendering described in Sec.~\ref{se:inverse_rendering} to synthesize photo-realistic images at large scale, which well addresses the shortcoming of the synthetic face images generated in~\cite{zhu2016face,richardson2016learning}. In particular, we choose 4000 face images (dataset A), in which faces are not occluded, from 300W~\cite{sagonas2013300} and Multi-pie~\cite{gross2010multi}. For each of the 4000 images, we use our optimization based inverse rendering method to obtain the parameter set $\chi$. Then, to make our coarse-layer network robust to expression and pose, we render new face images by randomly changing the pose parameters $\chi_{p}$ and the expression parameter $\bfalpha_{\txexp}$, each of which leads to a new parameter set $\widetilde{\chi}$. By doing the rasterization with $\widetilde{\chi}$, we can obtain the normals of all pixels in the new face region as described in Sec.~\ref{se:rendering}. With these normals and the albedos obtained according to Sec.~\ref{se:albedo_blend}, a new face is then rendered using Eq.~\eqref{eq:irradiance}. We also warp the background region of the source image to fit the new face region by using the image meshing~\cite{zhu2016face}. Fig.~\ref{fig:coarsedata} shows an example of generating three synthetic images from an input real images by simultaneously changing the expression and pose parameters. In this way, we generate a synthetic dataset of totally 80,000 face images for the Single-image CoarseNet training by randomly varying the expression and the pose parameters 20 times for each of the 4000 real face images.

\subsection{Single-image CoarseNet}
The input to our Single-image CoarseNet is a face image, and the output is the parameters related to the shape of 3D face and the projection, i.e. $\mathcal{T} = \{\alpha_{\textrm{id}}, \alpha_{\textrm{exp}}, s, pitch, yaw, roll, t_{x}, t_{y}\}$. The network is based on the Resnet-18~\cite{he2016deep} with the modification of changing the output number of the fully-connected layer to 185 (100 for identity, 79 for expression, 3 for rotation, 2 for translation and 1 for scale). The input image size is $224\times 224$.

As pointed out in~\cite{zhu2016face}, different parameters in $\mathcal{T}$ have different influence to the estimated geometry. Direct MSE (mean square error) loss on $\mathcal{T}$ might not lead to good geometry reconstruction. \cite{zhu2016face} uses a weighted MSE loss, where the weights are based on the projected vertex distances. \cite{richardson2016learning} uses 3D vertex distances to measure the loss from the geometry parameters and MSE for the pose parameters. Considering these vertex based distance measures are calculated on the vertex grid, which might not well measure how the parameters fit the input face image, in this work we use a loss function that computes the distance between the ground-truth parameters $\mathcal{T}_{g}$ and the network output parameters $\mathcal{T}_{n}$ at the per-pixel level.

In particular, we first do the rasterization with the ground-truth parameters $\mathcal{T}_{g}$ to get the underlying triangle index and the barycentric coordinates for each pixel in the face region. With this information, we then construct the pixels' 3D average $\bar{\bfp}_{q}$, base $\bfA_{q, \txid}$ and base $\bfA_{q, \txexp}$ by barycentrically interpolating the corresponding rows in $\bar{\bfp}$, $\bfA_{\txid}$, $\bfA_{\txexp}$, respectively. In this way, given parameters $\mathcal{T}$, we can project all the corresponding 3D locations of the pixels onto the image plane using
\begin{equation}\label{eq:Tk}
Proj(\mathcal{T}) = \Pi R (\bar{\bfp}_{q} + \bfA_{q, \txid}\bfalpha_{\txid} + \bfA_{q,\txexp} \bfalpha_{\txexp}) + \bft.
\end{equation}
Then the loss between the ground-truth parameters $\mathcal{T}_{g}$ and the network output parameters $\mathcal{T}_{n}$ is defined as:
\begin{equation}\label{eq:DTk}
\mathcal{D} (\mathcal{T}_{g}, \mathcal{T}_{n}) = \|Proj(\mathcal{T}_{g}) - Proj(\mathcal{T}_{n})\|_{2}^2.
\end{equation}
Note that there is no need to compute $Proj(\mathcal{T}_{g})$ since it corresponds to the original pixel locations in the image plane.

For better convergence, we further separate the loss in Eq.~\eqref{eq:DTk} into the pose-dependent loss as
\begin{equation}\label{eq:Lpose}
\mathcal{L}_{\textrm{pose}} = \|Proj(\mathcal{T}_{g}) - Proj(\mathcal{T}_{n,\textrm{pose}},\mathcal{T}_{g,\textrm{geo}})\|_{2}^2,
\end{equation}
where $\mathcal{T}_{\textrm{pose}} = \chi_{p}\cup \chi_{t}$ represents the pose parameters, and the geometry-dependent loss as
\begin{equation}\label{eq:Lgeo}
\mathcal{L}_{\textrm{geo}} = \|Proj(\mathcal{T}_{g}) - Proj(\mathcal{T}_{n,\textrm{geo}},\mathcal{T}_{g,\textrm{pose}})\|_{2}^2,
\end{equation}
where $\mathcal{T}_{\textrm{geo}} = \chi_{g}$ represents the geometry parameters. In Eq.~\eqref{eq:Lpose} and Eq.~\eqref{eq:Lgeo}, $Proj(\mathcal{T}_{n,\textrm{pose}},\mathcal{T}_{g,\textrm{geo}})$
(resp., $Proj(\mathcal{T}_{n,\textrm{geo}},\mathcal{T}_{g,\textrm{pose}})$) refers to the projection with the ground-truth geometry (resp., pose) parameters and the network estimated pose (resp., geometry) parameters.

The final loss is a weighted sum of the two losses:
\begin{equation}
\mathcal{L} = w \cdot \mathcal{L}_{\textrm{pose}} + (1-w) \cdot \mathcal{L}_{\textrm{geo}},
\label{eq:totalloss}
\end{equation}
where $w$ is the tradeoff parameter. We set $w = \frac{\mathcal{L}_{\textrm{geo}}}{\mathcal{L}_{\textrm{pose}}+\mathcal{L}_{\textrm{geo}}}$ for balancing the two losses and we assume $w$ is a constant when computing the derivatives for back propagation.

\section{Tracking CoarseNet for Coarse Reconstruction from Monocular Video}
The purpose of Tracking CoarseNet is to predict the current frame's parameters, given not only the current video frame but also the previous frame's parameters. As there does not exist large-scale dataset that captures the correlations among adjacent video frames, our
Tracking CoarseNet also faces the problem of no sufficient well-labelled training
data. Similarly, we synthesize training data for Tracking CoarseNet. However, it is non-trivial to reuse the ($k - 1$)-th frame's parameters to predict $k$-th frame's parameters. Directly using all the previous frame's parameters as the input to Tracking CoarseNet will introduce too many uncertainties during training, which results in huge complexity in synthesizing adjacent video frames for training, and make the training hard to converge and the testing unstable. Through vast experiments, we find that only utilizing the previous frame's pose parameters is a good way to inherit the coherence while keeping the network trainable and stable.

Specifically, the input to the tracking network is the $k$-th face frame cropped by the $k-1$ frame's landmarks and a Projected Normalized Coordinate Code (PNCC)~\cite{zhu2016face} rendered using the $k-1$ frame's pose parameters $\chi_{p}^{k-1}$, $\chi_{t}^{k-1}$ and the mean 3D face $\bar{\bfp}$ in Eq.~\eqref{eq:3dmm_geo}. The output of the tracking network is parameters $\mathcal{T}^k = \{\bfalpha_{\txid}^k, \bfalpha_{\txexp}^k, \bfalpha_{\txalb}^k, \delta^k(s), \delta^k(pitch), \delta^k(yaw), \delta^k(roll), \delta^k(\bft), \bfr^k\}$, where $\delta(\cdot)$ denotes the difference between the current frame and the previous frame. Note that here the output also includes albedo and lighting parameters, which could be used for different video editing applications.

The network structure is the same as Single-image CoarseNet except that the output number of the fully-connected layer is 312 (100 for identity, 79 for expression, 3 for rotation, 2 for translation, 1 for scale, 100 for albedo and 27 for lighting coefficients). In addition to the loss terms $\mathcal{L}_{\textrm{pose}}$ and $\mathcal{L}_{\textrm{geo}}$ defined in Eq.~\eqref{eq:Lpose} and Eq.~\eqref{eq:Lgeo} respectively, Tracking CoarseNet also uses another term for $\bfalpha^k_{\txalb}$ and $\bfr^k$ that measures the distance between the rendered image and the input frame:
\begin{equation}
\mathcal{L}_{\textrm{col}} = \|I^k_{\textrm{ren}}(\bfalpha^k_{\txalb}, \bfr^k) - I^k_{\textrm{in}}\|_{2}^2,
\end{equation}
where $I^k_{\textrm{ren}}(\bfalpha^k_{\txalb}, \bfr^k)$ is the rendered face image with the groundtruth geometry and pose, and the estimated albedo and lighting, and $I^k_{\textrm{in}}$ is the input face frame. In this way, the final total loss becomes a weighted sum of the three losses:
\begin{equation}\label{eq:L}
\mathcal{L} = w_1 \cdot \mathcal{L}_{\textrm{pose}} + w_2 \cdot \mathcal{L}_{\textrm{geo}} + (1-w_1-w_2) \cdot \mathcal{L}_{\textrm{col}},
\end{equation}
where $w_1 = \frac{\mathcal{L}_{\textrm{geo}} + \mathcal{L}_{\textrm{col}}}{2(\mathcal{L}_{\textrm{pose}} + \mathcal{L}_{\textrm{geo}} + \mathcal{L}_{\textrm{col}})}$ and $w_2 = \frac{\mathcal{L}_{\textrm{pos}} + \mathcal{L}_{\textrm{col}}}{2(\mathcal{L}_{\textrm{pose}} + \mathcal{L}_{\textrm{geo}} + \mathcal{L}_{\textrm{col}})}$ are the tradeoff parameters to balance the three losses, and we assume $w_1$ and $w_2$ are constant when computing the derivatives for back propagation.

\begin{figure}
	\centering
	\includegraphics[width=1\columnwidth]{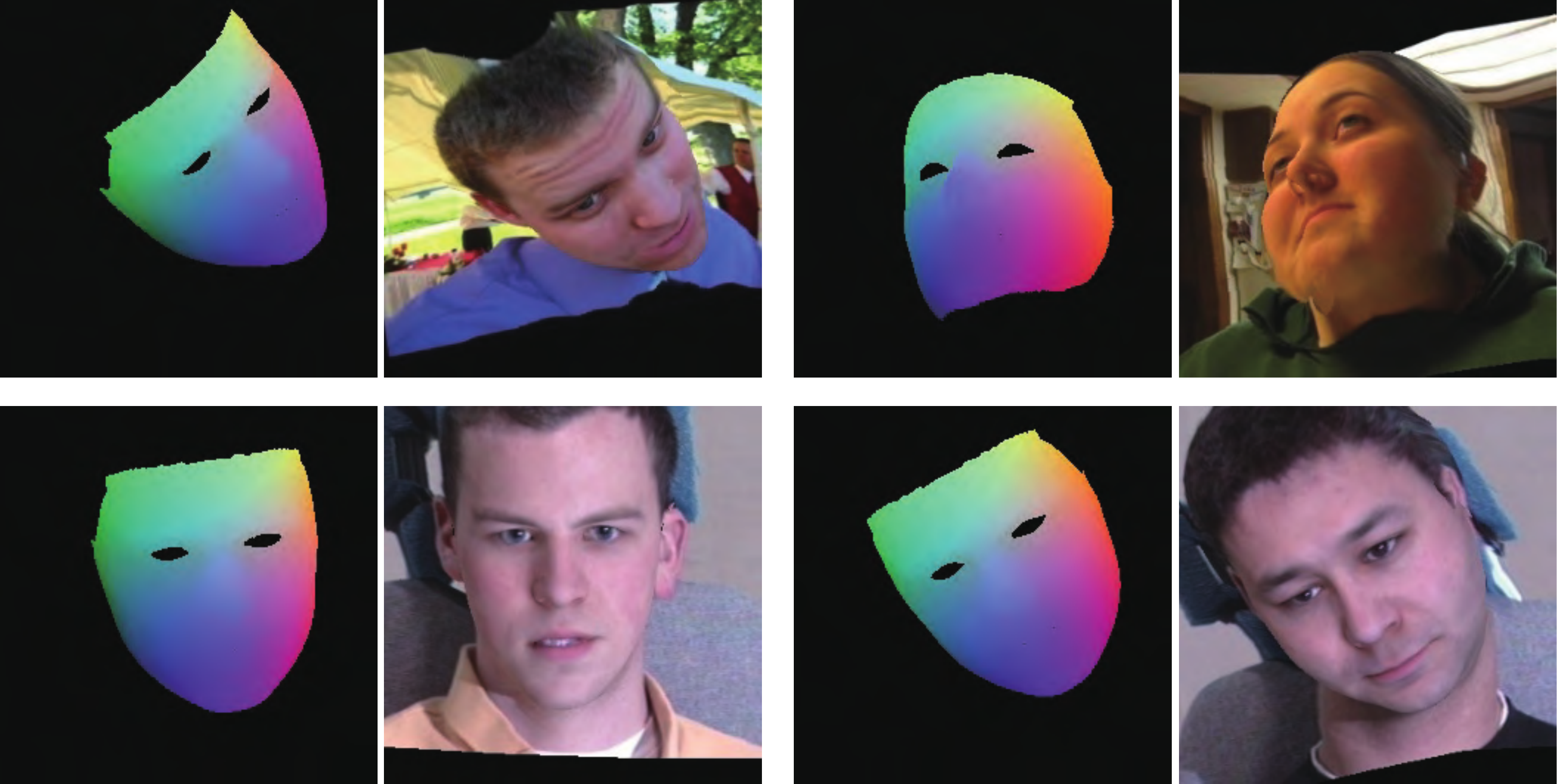}
	\caption{Examples of adjacent frame simulations. For each pair, the left is the PNCC image generated by simulating the previous frame, and the right is the current face frame.}
	\label{fig:pair}
\end{figure}

\para{Training data generation for Tracking CoarseNet} To train Tracking CoarseNet, large-scale adjacent video frame pairs with ground-truth parameters $\chi$ are needed as training data. Again, there is no such public dataset. To address this problem, we propose to simulate adjacent video frames, i.e., to generate the previous frame for each of the 80,000 synthesized images used in the Single-image CoarseNet training. Randomly varying the parameter set $\widetilde{\chi}$ for a training image does not capture the tight correlations among adjacent frames. Thus, we propose to do simulation by analysing the distribution of the previous frame's parameters $\chi^{k-1}$ given the current $k$-th frame from real videos. Considering our tracking network only makes use of the previous frame's pose parameters, we just need to obtain the distribution of $\chi^{k-1}_p$ and $\chi^{k-1}_t$ given $\chi^{k}_p$ and $\chi^{k}_t$. Particularly, we assume each parameter in $\delta^k(\chi_{p})= \chi^{k-1}_p - \chi^{k}_p$ and $\delta^k(\chi_{t})= \chi^{k-1}_t - \chi^{k}_t$ follows normal distribution. We extract about 160,000 adjacent frame pairs from the 300-VW video dataset~\cite{shen2015first} and use our Single-image CoarseNet to get the parameters for fitting the normal distribution. Finally, for each of the 80,000 synthesized images, we can simulate its previous frame by generating $\widetilde{\chi}^{k-1}_p$ and $\widetilde{\chi}^{k-1}_t$ according to the obtained normal distribution. Examples of several simulated pairs with the previous frame's PNCC and the current image are shown in Fig.~\ref{fig:pair}.

\section{FineNet for Fine-scale Geometry Reconstruction}
\label{sect:fine}
In this section, we present our solution on how to train a fine-layer CNN (called FineNet). The input to FineNet is a coarse depth map stacked with the face image. The coarse depth map is generated by using the method described in Sec.~\ref{se:geo_refine} with the parameters $\mathcal{T}$ estimated by either Single-image CoarseNet or Tracking CoarseNet. The output of our FineNet is a per-pixel displacement map. Again, the key challenge here is that there is no fine-scale face dataset available that can provide a large number of detailed face geometries with their corresponding 2D images, as pointed out in~\cite{richardson2016learning}. In addition, the existing morphable face models such as 3DMM cannot capture the fine-scale face details. \cite{richardson2016learning} bypasses this challenge by converting the problem into an unsupervised setting, i.e. relating the output depth map to the 2D image by using the shading energy as the loss function. However, to make the back-propagation trackable under the shading energy, they have to use first-order spherical harmonics to model the lighting, which is not accurate.

In our work, instead of doing unsupervised training~\cite{richardson2016learning}, we go for fully supervised training of FineNet, i.e. directly constructing a large-scale detailed face dataset based on our developed inverse rendering and a novel face detail transfer approach, which will be elaborated below. Note that our FineNet architecture is based on the U-Net~\cite{ronneberger2015u} and we use Euclidean distance as the loss function.

\subsection{Constructing FineNet Training Data}
\begin{figure}
  \centering
\includegraphics[width=0.5\textwidth]{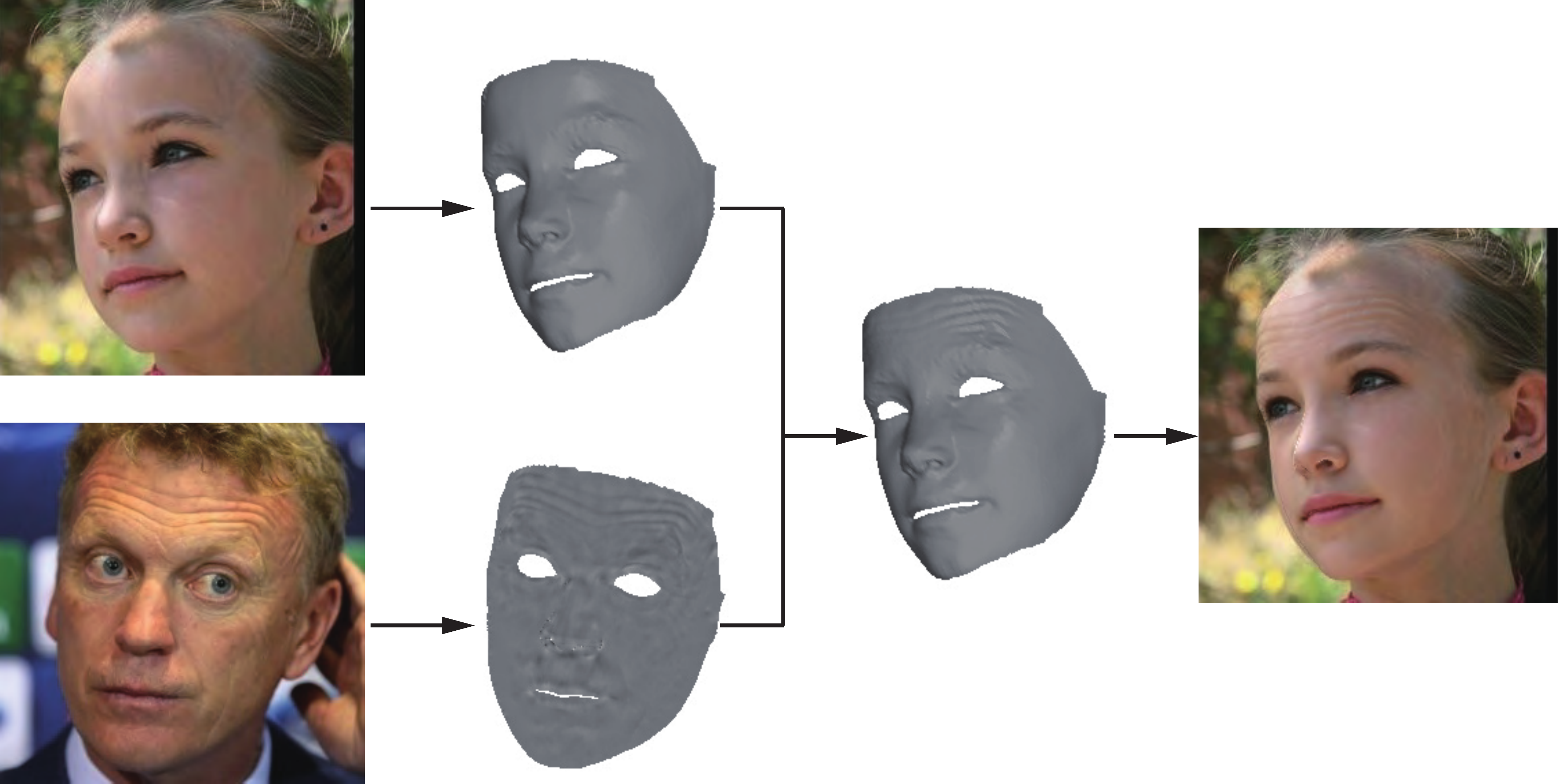}
  \caption{Synthetic data generation for training FineNet. Given a target face image without many geometry details (top left) and a source face image (bottom left) that is rich of wrinkles, we first apply our developed inverse rendering on both images to obtain the projected geometry for target face (top second) and a displacement map for the source face (bottom right). Then we transfer the displacement map of the source face to the geometry of the target face. Finally we render the updated geometry to get a new face image (top right) which contains the same type of wrinkles as the source face.}\label{fig:finedata}
\end{figure}

Our synthesized training data for FineNet is generated by transferring the displacement map from a source face image with fine-scale details such as wrinkles and folds to other target face images without the details. Fig.~\ref{fig:finedata} gives such an example. In particular, we first apply our developed inverse rendering in Sec.~\ref{se:inverse_rendering} on both images. Then we find correspondences between the source image pixels and the target image pixels using the rasterization information described in Sec.~\ref{se:rendering}. That is, for a pixel $(i, j)$ in the target face region, if its underlying triangle is visible in the source image, we find its corresponding 3D location on the target 3D mesh by barycentric interpolation, and then we project the 3D location onto the source image plane using Eq.~\eqref{eq:project} to get the corresponding pixel $(i', j')$. With these correspondences, the original source displacement $\bm{d}_{s}$ and the original target displacement $\bm{d}_{t}$, a new displacement $\widetilde{\bm{d}_t}$ for the target image is generated by matching its gradients with the scaled source displacement gradient in the intersected region $\Omega$ by solving the following poisson problem:
\begin{align}
	\min_{\widetilde{\bfd}_t}~ &~ \sum\limits_{(i, j)\in \Omega} \|\nabla \widetilde{\bfd}_t(i, j) - \bfw(i, j)\|^{2},\nonumber\\
	\textrm{s.t.}~ &~ \widetilde{\bfd}_t(i, j) = \bfd_{t}(i, j) \qquad (i, j) \in \partial\Omega
\label{eq:detailPoisson}
\end{align}
where $\bfw(i, j) = s_d[\bfd_{s}(i' + 1, j') - \bfd_{s}(i', j'), \bfd_{s}(i', j' + 1) - \bfd_{s}(i', j')]^{T}$ and $s_d$ is a scale factor within the range $[0.7, 1.3]$ so as to create different displacement fields. After that, we add $\widetilde{\bfd}_t$ into the coarse target depth $\bfz$ to get the final depth map. Then the normals of the target face pixels are updated as in Sec.~\ref{se:geo_refine}. With the updated normals, a new face image is rendered using Eq.~\eqref{eq:irradiance}.

We would like to point out that besides generating a large number of detailed face images to train the network, there are also other benefits to do such detail transfer. First, by rendering the same type of detail information under different lighting conditions, we can train our FineNet to be robust to lighting. Second, by changing the scale of the displacement randomly, our method can be trained to be robust to different scales of details.

For the details of the dataset construction, we first download 1000 real face images (dataset B) that contain rich geometry details from internet. Then, we transfer the details from dataset B to the 4000 real face images in dataset A, the one used in constructing synthetic data for Single-image CoarseNet. For every image in A, we randomly choose 30 images in B for transferring. In this way, we construct a synthesized fine-detailed face image dataset of totally 120,000 images.

\section{Experiments}
In this section, we conduct qualitative and quantitative evaluation on the proposed detailed 3D face reconstruction and tracking framework and compare it with the state-of-the-art methods.

\para{Experimental setup and runtime} We train the CNNs via the CAFFE~\cite{jia2014caffe} framework. Single-image CoarseNet takes the input of a color face image with size $224\times 224\times 3$, and Tracking CoarseNet and FineNet respectively take the inputs of $256\times 256\times 6$ (a color image and a PNCC) and $256\times 256\times 2$ (a gray image and its coarse depth). We train all the networks using Adam solver with the mini-batch size of 100 and 30k iterations. The base learning rate is set to be $0.00005$.

The CNN based 3D face reconstruction and tracking are implemented in C++ and tested on various face images and videos. All experiments were conducted on a desktop PC with a quad-core Intel CPU i7, 4GB RAM and NVIDIA GTX 1070 GPU. As for the running time for each frame, it takes 5 ms for CoarseNet and 15 ms for FineNet.

\begin{figure}
	\centering
	\includegraphics[width=1\columnwidth]{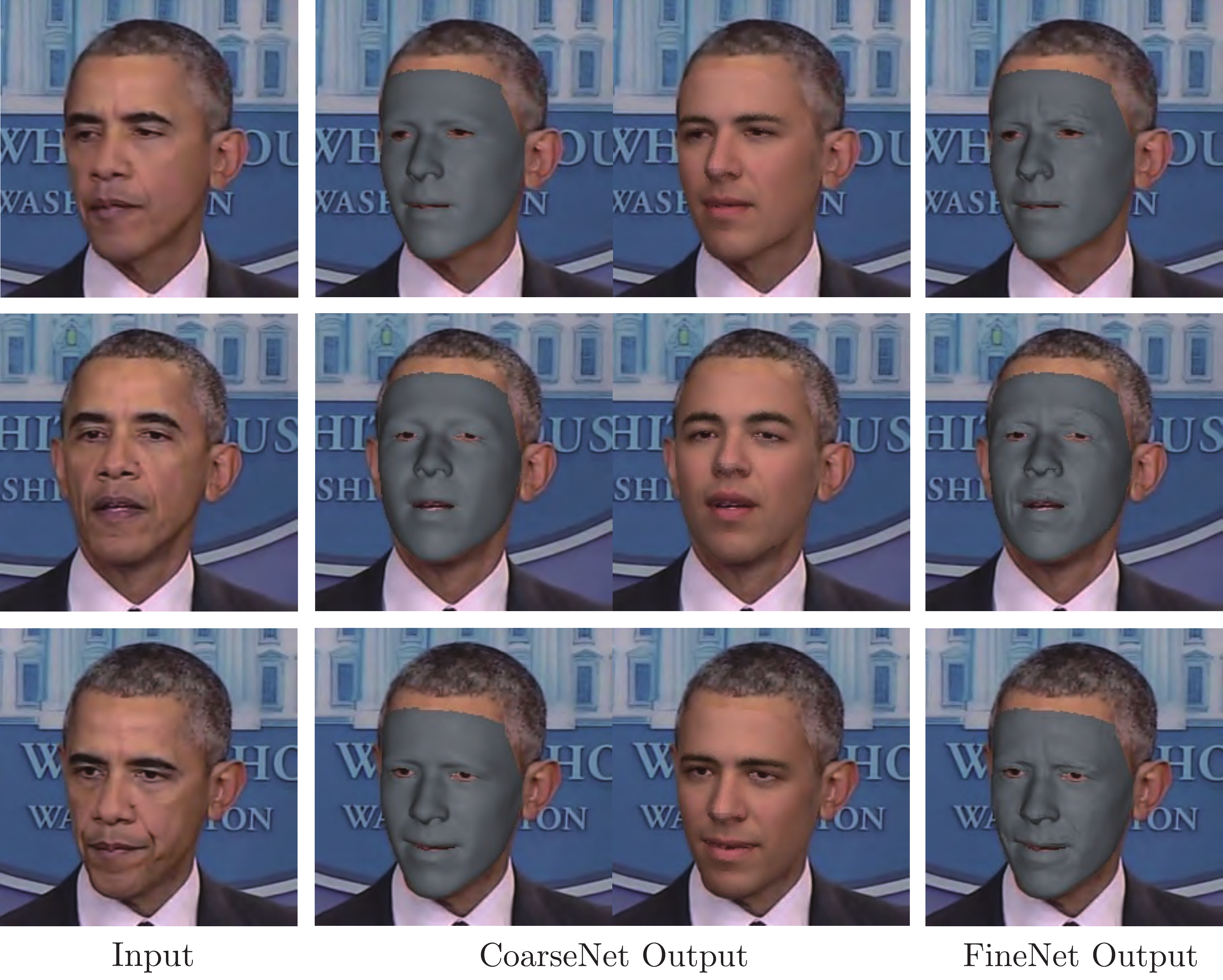}
	\caption{Results of our two-stage CNN based face tracking. Left: four frames of a video. Middle: results of Tracking CoarseNet (projected mesh and rendered face). Right: results of FineNet.}
	\label{fig:multi_layer}
\end{figure}

\subsection{Results of Dense 3D Face Reconstruction from Monocular Video}

\para{CoarseNet vs. FineNet} Our approach is to progressively and continuously estimate the detailed facial geometry, albedo and lighting parameters from a monocular face video. Fig.~\ref{fig:multi_layer} shows the tracking output results of the two stages. The results of CoarseNet include the smooth geometry and the corresponding rendered face image shown in the middle column. The FineNet further predicts the pixel level displacement given in the last column. We can see that CoarseNet produces smooth geometry and well matched rendered face images, which show the good recovery of pose, albedo, lighting and projection parameters, and FineNet nicely recovers the geometry details such as wrinkles. A complete reconstruction results of all the video frames are given in the accompanying video or via the link: \url{https://youtu.be/dghlMXxD-rk}.

\begin{figure}
	\centering
	\includegraphics[width=1\columnwidth]{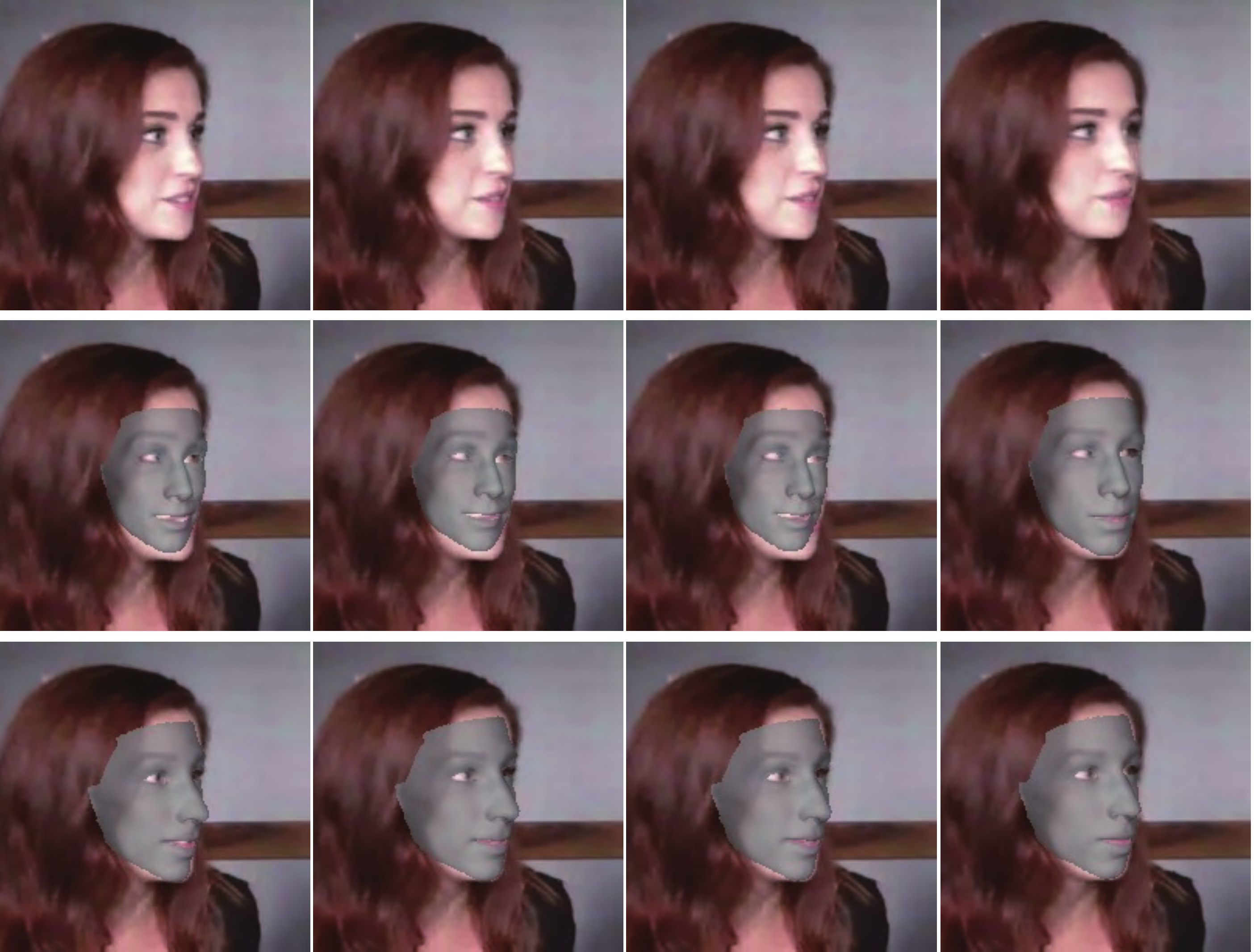}
	\caption{Comparisons with image-based dense face tracking. Top row: four continuous frames from a video. Middle row: results of using our Single-image CoarseNet on each frame. Bottom row: results of our Tracking CoarseNet. It can be observed that Tracking CoarseNet achieves more robust tracking.}
	\label{fig:compare_fist}
\end{figure}

\para{Single-image CoarseNet vs. Tracking CoarseNet} Given a RGB video, a straightforward way for dense face tracking is to treat all frames as independent face images, and apply our Single-image CoarseNet on each frame, followed by applying FineNet. Thus, we give a comparison of our proposed Tracking CoarseNet, which estimates the differences of the pose parameters w.r.t. the previous frame, with the baseline that simply uses our Single-image CoarseNet on each frame. As demonstrated in Fig.~\ref{fig:compare_fist}, Tracking CoarseNet achieves more robust tracking than the baseline, since it well utilizes the guidance from the previous frame's pose.

\begin{figure}
	\begin{center}
	\includegraphics[width=1\columnwidth]{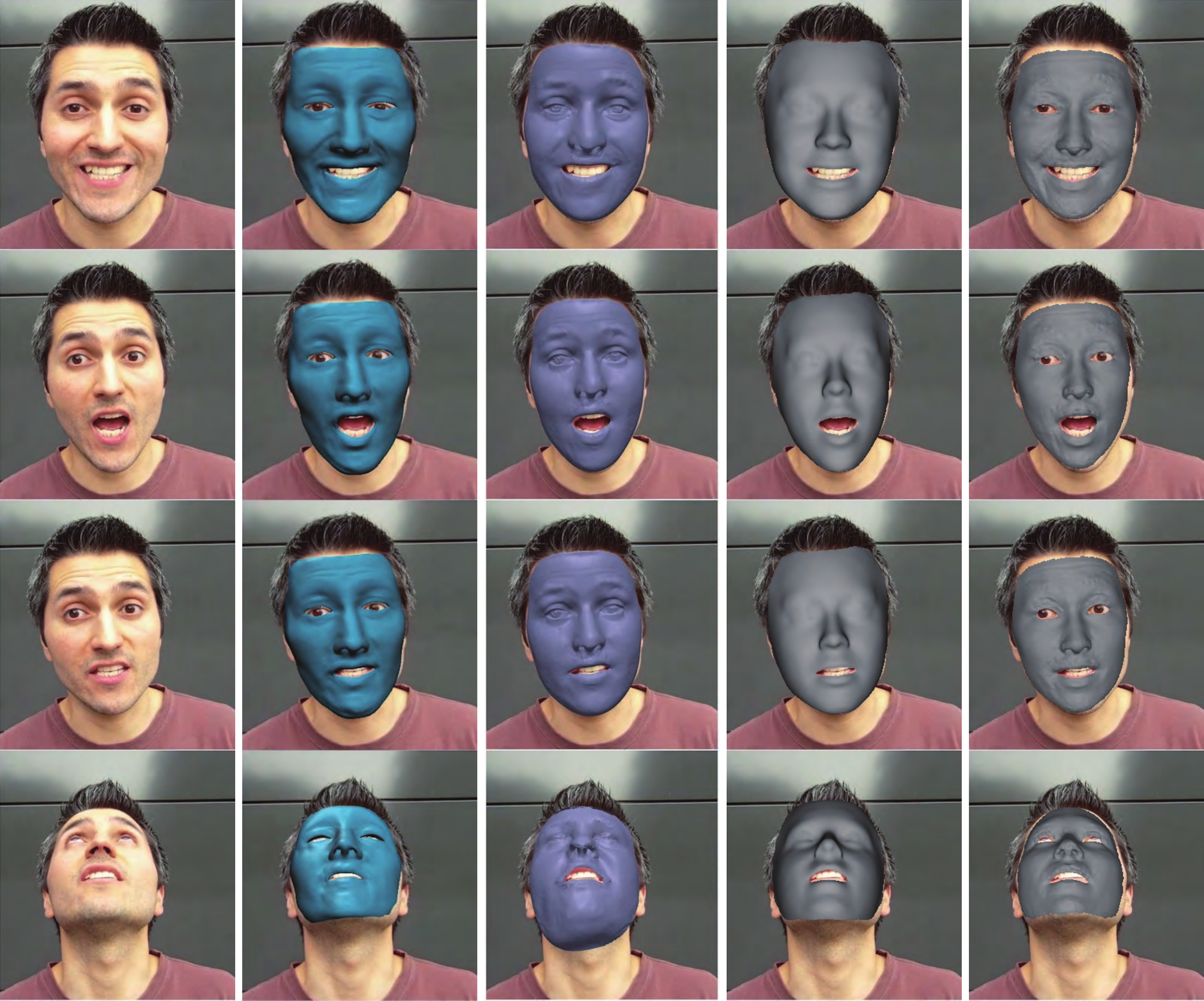}
	\end{center}
	\qquad Input \quad \  ~\cite{shi2014automatic}(-) \quad \cite{garrido2016reconstruction}(175.5s) ~\cite{huber2016multiresolution}(100ms) \ Ours(25ms)
	\caption{\edit{Comparisons with the state-of-art dense face tracking methods~\cite{shi2014automatic,garrido2016reconstruction,huber2016multiresolution}.} \editnew{The average computation time for each frame is given in the bracket. \cite{shi2014automatic} does not report the running time of their method, while it should take much longer time than ours since it iteratively solves several complex optimization problems.}}
	\label{fig:compare_garri}
\end{figure}

\begin{figure}
  \includegraphics[width=0.5\textwidth]{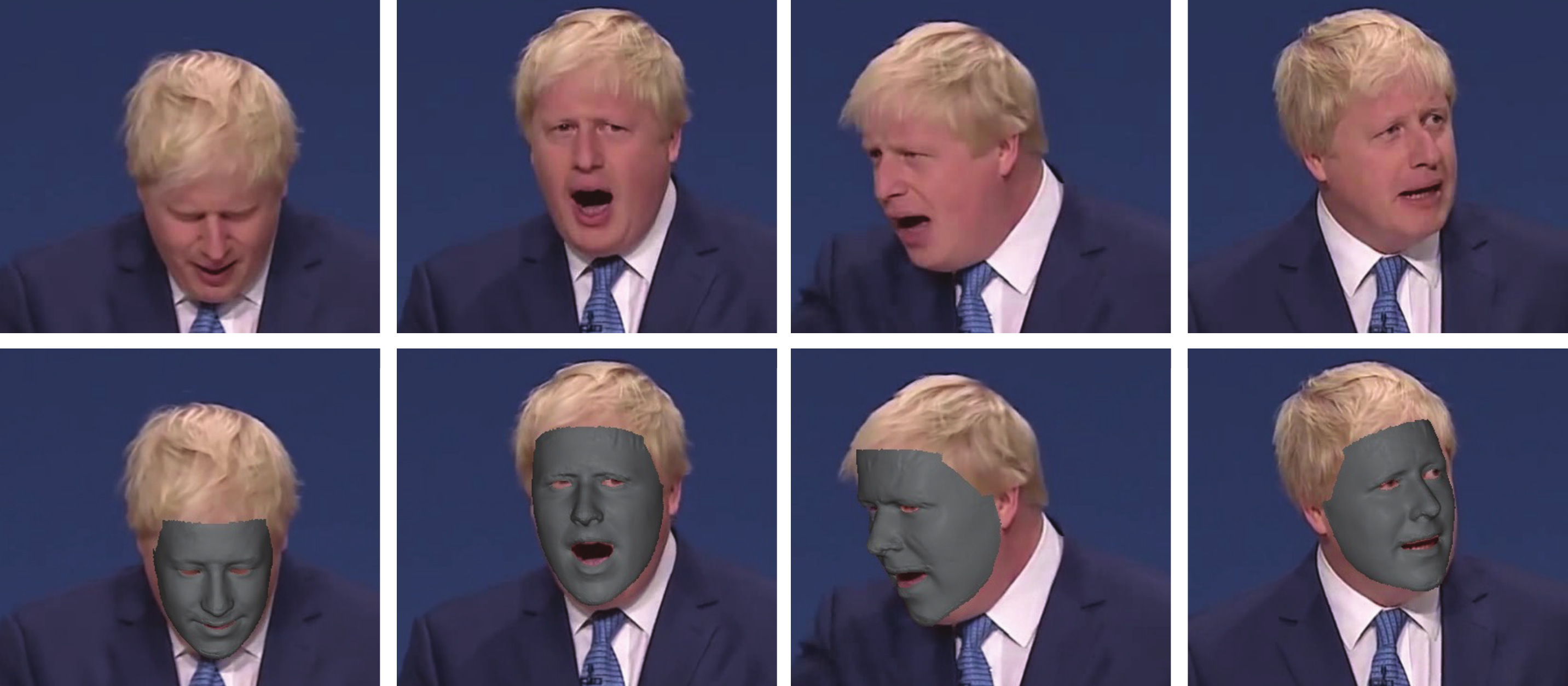}
  \caption{Reconstruction results for faces with large poses and extreme expressions. Top row: several frames from one input video. Bottom row: the reconstructed face shapes with geometry details. See the complete sequence in the accompanying video or via the link: \url{https://youtu.be/dghlMXxD-rk}.
}
  \label{fig:largepose}
\end{figure}

\para{Comparisons with dense face tracking methods}
We compare our method with the state-of-the-art monocular video based dense face tracking methods~\cite{shi2014automatic,garrido2016reconstruction,huber2016multiresolution}.  ~\cite{shi2014automatic} performs 3D face reconstruction in an iterative manner. In each iteration, they first reconstruct coarse-scale facial geometry from sparse facial features and then refine the geometry via shape from shading. ~\cite{garrido2016reconstruction} employs a multi-layer approach to reconstruct fine-scale details. They encode different scales of 3D face geometry on three different layers and do optimization for each layer. \edit{~\cite{huber2016multiresolution} reconstructs the 3D face shape by only fitting the 2D landmarks via 3DMM, and we can observe that \cite{huber2016multiresolution} can only produce smooth face reconstruction.} \editnew{As shown in Fig.~\ref{fig:compare_garri}, our method produces visually better results compared to \cite{huber2016multiresolution}, and comparable results compared to \cite{garrido2016reconstruction} and \cite{shi2014automatic}.

Different from optimization based methods, our learning based approach is much faster while obtaining comparable or better results.} Our method is several orders of magnitude faster than the state-of-the-art optimization-based approach~\cite{garrido2016reconstruction}, i.e., 5 ms for CoarseNet and 15 ms for FineNet with our hardware setting, while 175.5s reported in their paper~\cite{garrido2016reconstruction}. It needs to be pointed out that the existing optimization based dense tracking methods need facial landmark constraints. Therefore, they might not reconstruct well for faces with large poses and extreme expressions. On the other hand, we do large-scale photo-realistic image synthesis that includes many challenging data with well labelled parameters, and thus we can handle those challenging cases as demonstrated in Fig.~\ref{fig:largepose}.

\begin{figure}
  \centering
  \includegraphics[width=0.5\textwidth]{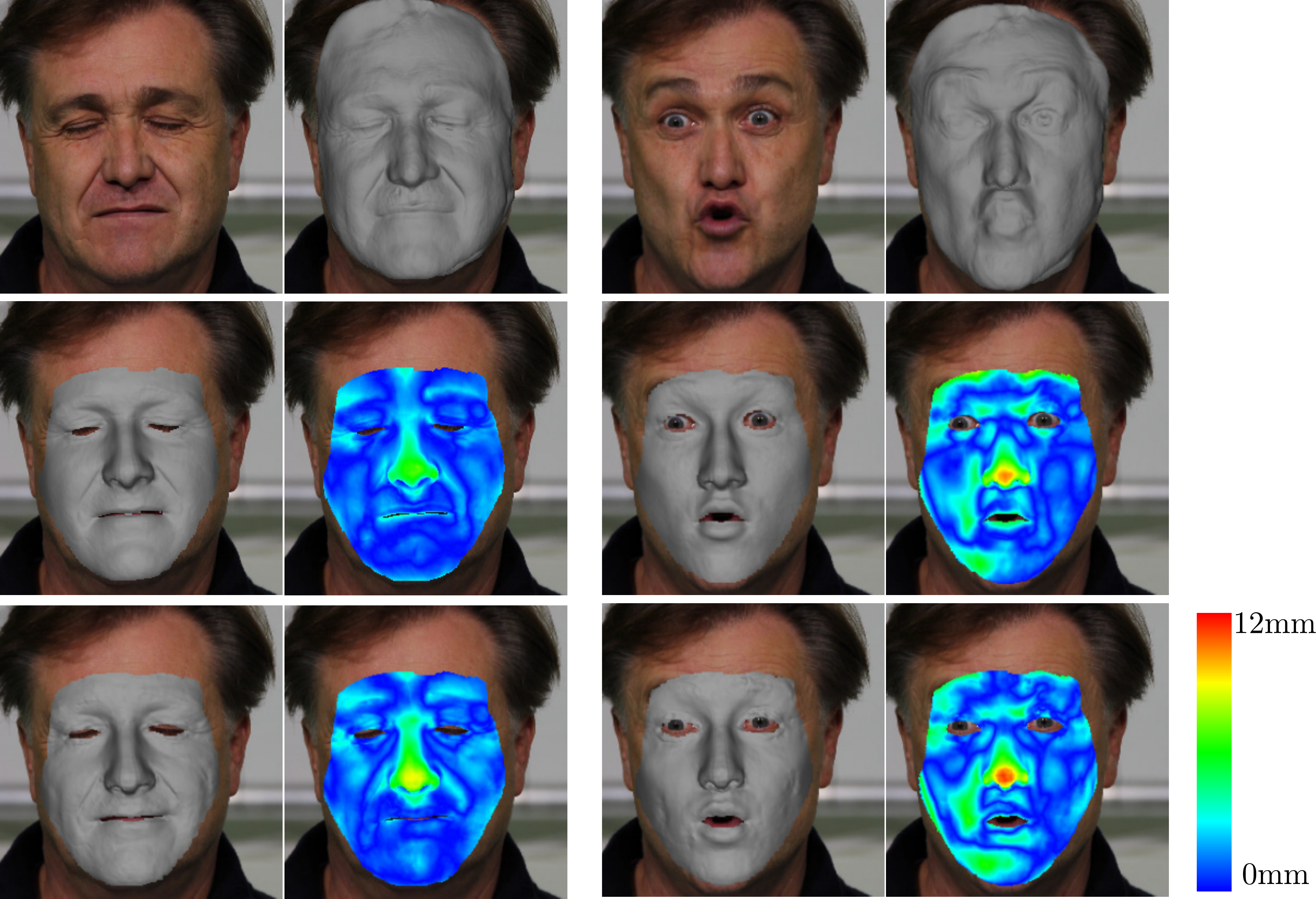}
  \caption{Comparisons of our inverse rendering and our learning based dense face tracking solution. From top to bottom: input face video frame and groundtruth mesh in dataset~\cite{ValgaertsWBST12}, results of the inverse rendering approach, results of our learning based dense face tracking solution.}
\label{fig:error_map}
\end{figure}

\begin{figure}
  \centering
  \includegraphics[width=0.5\textwidth]{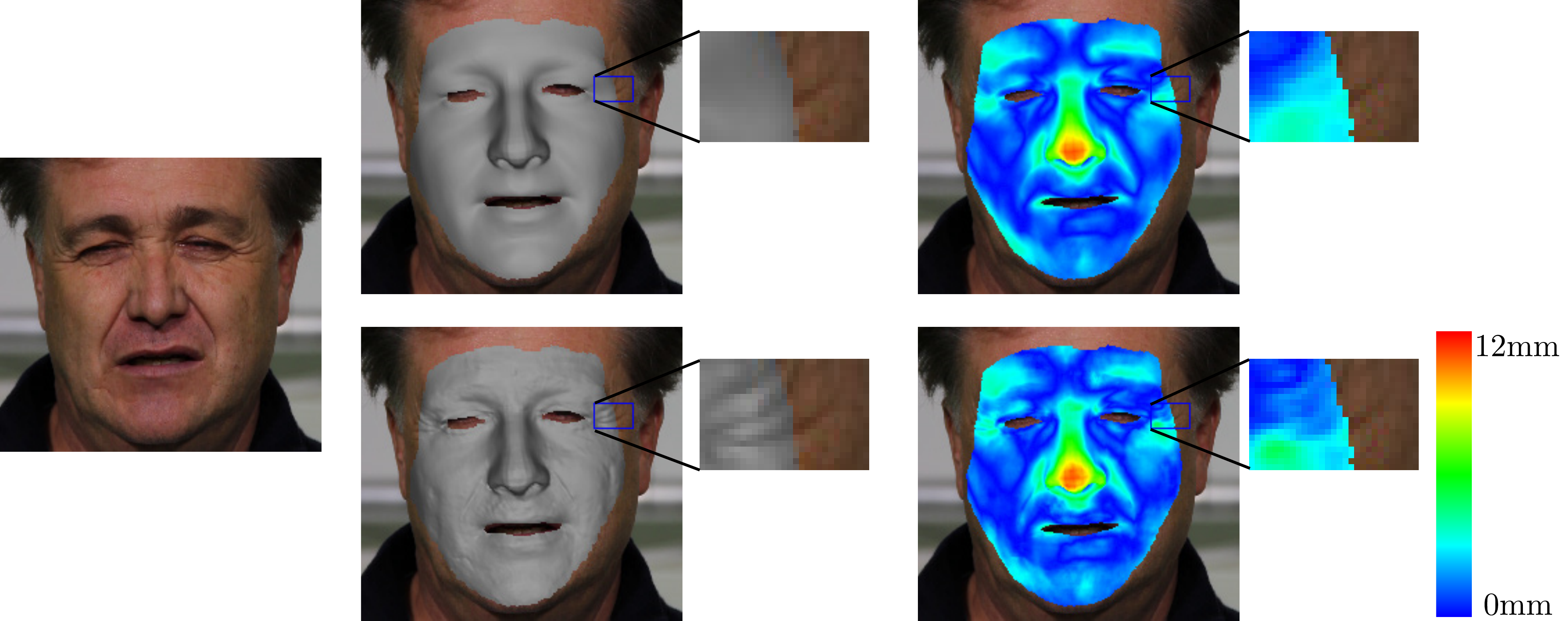}
  \caption{Comparison of our CoarseNet and FineNet on a small region that is rich of wrinkles. On the left is the input frame, on the top are results of CoarseNet, on the bottom are results of FineNet. On the subregion, the mean error is 2.20mm for CoarseNet and 2.03mm for FineNet.}
\label{fig:sub_error_map}
\end{figure}

\para{Quantitative results of face reconstruction from monocular video}
For quantitative evaluation, we test on the FaceCap dataset~\cite{ValgaertsWBST12}. The dataset consists of 200 frames along with 3D meshes constructed using the binocular approach. We compare our proposed inverse rendering approach and our learning based solutions including Tracking CoarseNet and Tracking CoarseNet+FineNet. For each method, we register the depth cloud to the groundtruth 3D mesh and compare point to point distance. Table~\ref{tab:quant_video} shows the average point-to-point distance results. It can be seen that our proposed inverse rendering achieves an average distance of 1.81mm, which is quite accurate. It demonstrates the suitability of using the inverse rendering results for constructing the training data. On the other hand, our CoarseNet+FineNet achieves an average distance of 2.08mm, which is comparable to that of the inverse rendering but with much faster processing speed (25ms vs 8s per frame). Some samples are shown in Fig.~\ref{fig:error_map}. \juyong{In addition, the reconstruction accuracy by CoarseNet+FineNet outperforms the one by CoarseNet alone. Since the face region containing wrinkles is only a small part of the whole face region, the difference is not significant since the accuracy statistics is computed over a large face region. By comparing the reconstruction accuracy on a small region that contains wrinkles, the improvement is more obvious, as shown in Fig.~\ref{fig:sub_error_map}.}

\begin{table}
 \centering
 \caption{Quantitative results of dense face reconstruction from monocular video.}  \label{tab:quant_video} \vspace{0.1in}
\begin{tabular}{|c|c|c|}
\hline
\multicolumn{3}{|c|}{Average point-to-point distance (mm)} \\ \hline
Inverse rendering   &CoarseNet  &CoarseNet+FineNet \\ \hline
1.81                &   2.11        &2.08 \\
\hline
\end{tabular}
\end{table}

\begin{figure}
	\centering
	\includegraphics[width=0.5\textwidth]{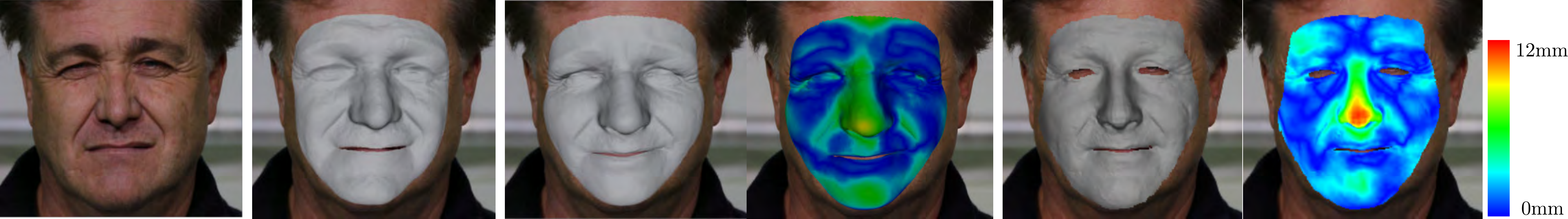}
	Input  \quad Stereo\cite{ValgaertsWBST12}   \qquad \quad \cite{garrido2016reconstruction}  \qquad  \qquad \qquad Ours  \qquad \ \
	\caption{The reconstruction accuracy comparison. The reconstruction quality of our dense face tracking method is comparable to the optimization based method~\cite{garrido2016reconstruction} but with much faster processing speed. The groundtruth mesh is constructed using the binocular approach~\cite{ValgaertsWBST12}.}
	\label{fig:geometryError}
\end{figure}

For the quantitative comparison with the state-of-the-art monocular video based face tracking method~\cite{garrido2016reconstruction}, we evaluate the geometric accuracy of the reconstruction of a video frame with rich face details (note that~\cite{garrido2016reconstruction} did not provide the results for the entire video). Fig.~\ref{fig:geometryError} shows the results, where our method achieves a mean error of 1.96mm compared to the groundtruth 3D face shape generated by the binocular facial performance capture proposed in~\cite{ValgaertsWBST12}. We can see that the result of our learning based face tracking method is quite close to the groundtruth, and is comparable (1.96mm vs. 1.8mm) to that of the complex optimization based approach~\cite{garrido2016reconstruction} but with much faster processing speed.

\subsection{Results of Dense 3D Face Reconstruction from A Single Image}
\begin{figure}
  \centering
  \includegraphics[width=0.115\textwidth]{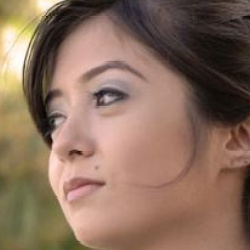}
  \includegraphics[width=0.115\textwidth]{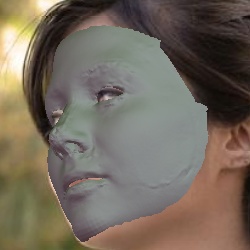}
  \includegraphics[width=0.115\textwidth]{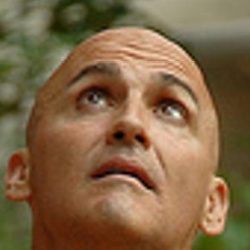}
  \includegraphics[width=0.115\textwidth]{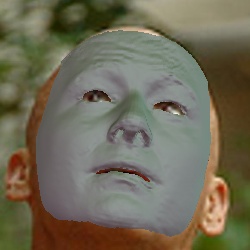}\\
\vspace{1.0mm}
    \includegraphics[width=0.115\textwidth]{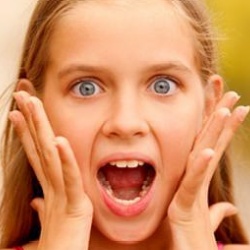}
  \includegraphics[width=0.115\textwidth]{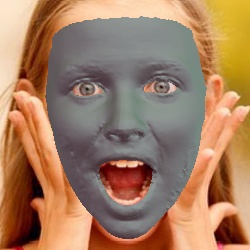}
  \includegraphics[width=0.115\textwidth]{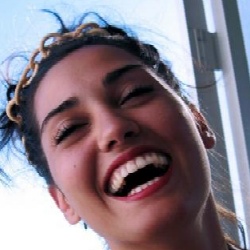}
  \includegraphics[width=0.115\textwidth]{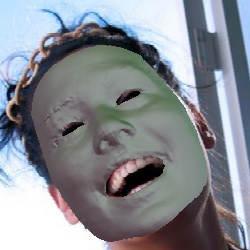}\\
\vspace{1.0mm}
    \includegraphics[width=0.115\textwidth]{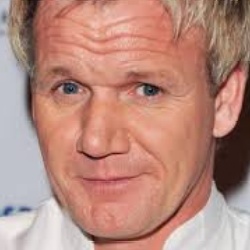}
  \includegraphics[width=0.115\textwidth]{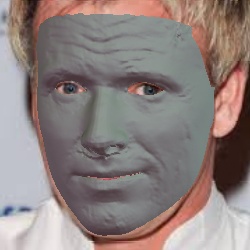}
  \includegraphics[width=0.115\textwidth]{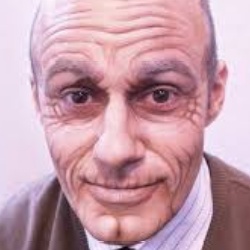}
  \includegraphics[width=0.115\textwidth]{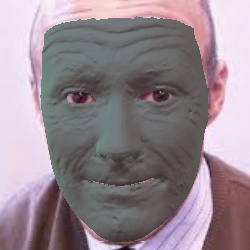}
  \caption{For each pair, on the left is the input face image; on the right is the projected 3D mesh reconstructed by our single-image based solution. The first, second and third rows respectively demonstrate that our method is robust to large poses, extreme expressions and different types and scales of wrinkles.}\label{fig:robustness}
\end{figure}

\para{Visual results of our single-image based reconstruction} To evaluate the single-image based reconstruction performance, we show the reconstruction results of our method (Single-image CoarseNet+FineNet) on some images from AFLW~\cite{kostinger2011annotated} dataset, VGG-Face dataset ~\cite{Parkhi15} and some face images downloaded from internet. The three rows in Fig.~\ref{fig:robustness} from top to bottom respectively show the projected 3D meshes reconstructed by our method under large poses, extreme expressions and face images with detailed wrinkles, which demonstrate that our method is robust to all of them.

\begin{figure}
\centering
  \includegraphics[width=0.15\textwidth]{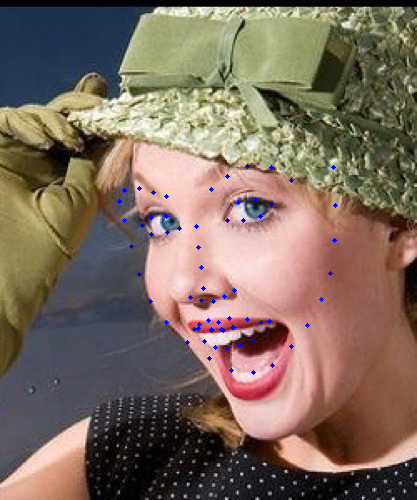}
  \includegraphics[width=0.15\textwidth]{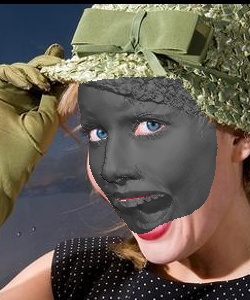}
  \includegraphics[width=0.15\textwidth]{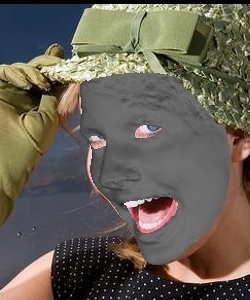}
\caption{From left to right: input face image with detected landmarks, geometry reconstructed by inverse rendering, geometry reconstructed by our learning based method. It can be seen that our inverse rendering approach fails to recover the face shape as the landmarks are not accurate. On the other hand, our proposed learning-based approach recovers the face shape well.}
\label{fig:optVSnet}
\end{figure}

\begin{figure}
  \centering
  \includegraphics[width=0.15\textwidth]{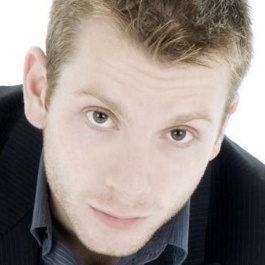}
  \includegraphics[width=0.15\textwidth]{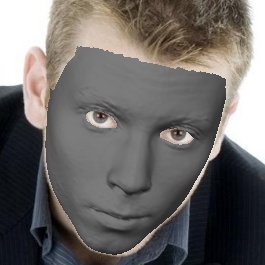}
  \includegraphics[width=0.15\textwidth]{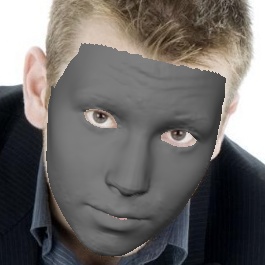}
  \caption{From left to right: input face image, geometry reconstructed by inverse rendering, geometry reconstructed by our learning based method. It can be seen that our method can better reconstruct unclear wrinkles under strong lighting.}\label{fig:lighting}
\end{figure}

\para{Comparisons with inverse rendering} Similar to the video input scenario, directly using our developed inverse rendering approach can also reconstruct detailed geometries from a single image, but our learning-based method does provide some advantages. First, unlike the inverse rendering approach, our learning-based method does not need face alignment information. Therefore, the learning-based method is more robust to input face image with large pose, as shown in Fig.~\ref{fig:optVSnet}. Second, once the two CNNs are trained, our learning method is much faster to reconstruct a face geometry from a single input image. Third, as we render the same type of wrinkles under different lightings and directly learn the geometry in a supervised manner, our method is more robust to lighting, as illustrated in Fig.~\ref{fig:lighting}. The  reason why the learning based method can do better in these scenarios lies in the large numbers of diverse training data we construct, which facilitate the learning of the two networks, while the inverse rendering approach only explores the information from each single image.

\begin{figure}
\includegraphics[width=0.5\textwidth]{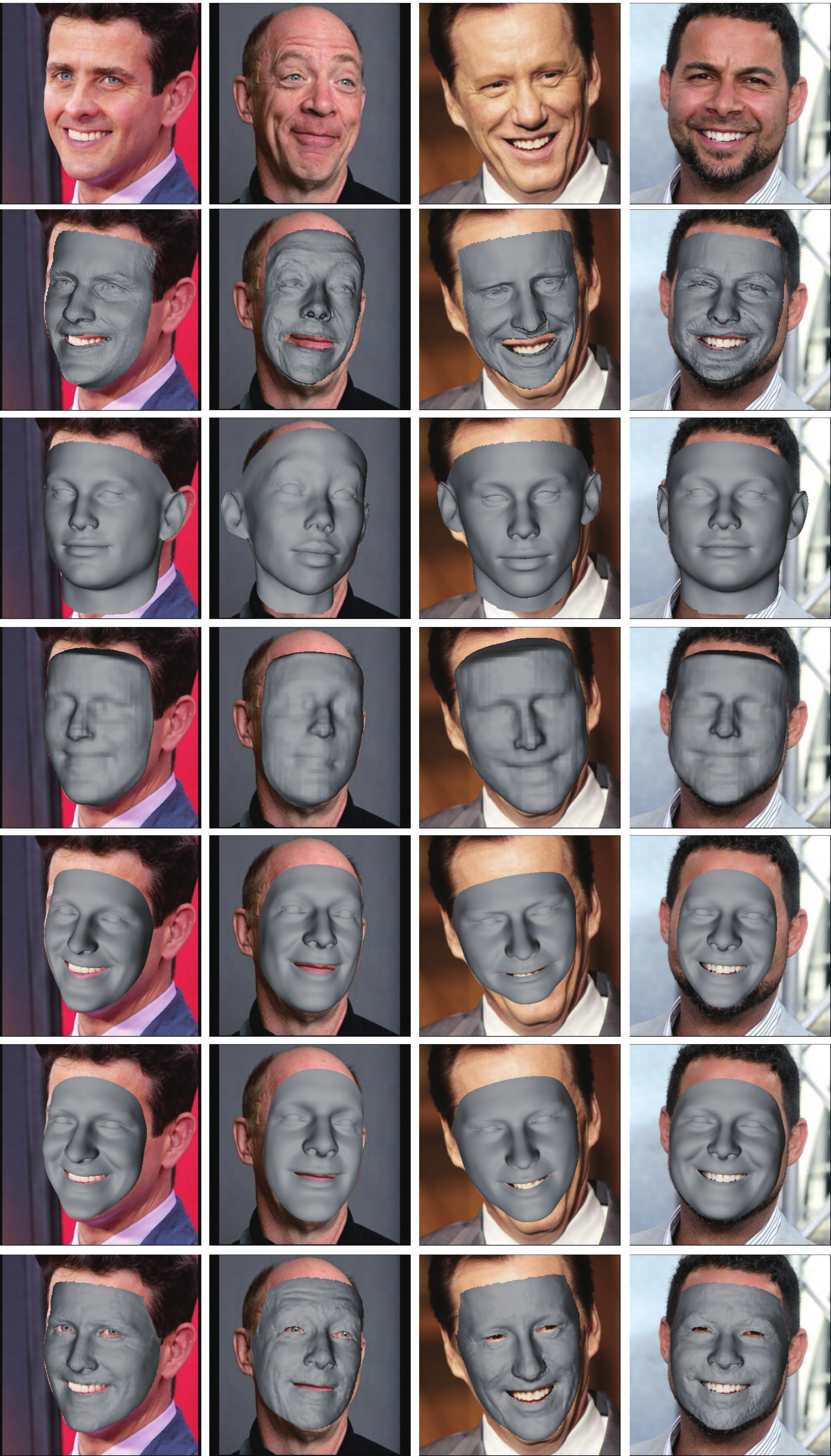}
\caption{\edit{Comparisons with the state-of-art methods. From the first row to the last row, it respectively shows the input images, and the results of~\cite{richardson2016learning}, \cite{bas2016fitting}, \cite{jackson2017large}, \cite{schonborn2017markov}, \cite{Egger2018} and ours. It can be seen that our results are more convincing in both the global geometry and the fine-scale details. \editnew{Note that the method of \cite{bas2016fitting} uses a 3DMM with identity variation only, and thus is not able to handle facial expressions well.}}}\label{fig:compare}
\end{figure}

\para{Comparisons with state-of-the-art single-image based face reconstruction}
\edit{We compare our method with~\cite{richardson2016learning,bas2016fitting,jackson2017large,schonborn2017markov,Egger2018} on single-image based face reconstruction. We thank the authors of~\cite{richardson2016learning} for providing us the same 11 images listed in~\cite{richardson2016learning}, as well as their results of another 8 images supplied by us. We show the reconstruction results of 4 images in Fig.~\ref{fig:compare} and the full comparisons on all the 19 images are given in the accompanying material. It can be observed that our method produces more convincing reconstruction results in both the global geometry (see the mouth regions) and the fine-scale details (see the forehead regions). The reconstruction results of the methods~\cite{bas2016fitting,jackson2017large,schonborn2017markov,Egger2018} are generated using the source codes provided by the authors\footnote{\url{https://github.com/waps101/3DMM_edges}}\footnote{\url{https://github.com/AaronJackson/vrn}}\footnote{\url{https://github.com/unibas-gravis/basel-face-pipeline}}\footnote{\url{https://github.com/unibas-gravis/scalismo-faces}}.}

The reasons why our method produces better results than~\cite{richardson2016learning} are threefold: 1) For CoarseNet training, ~\cite{richardson2016learning} only renders face region and uses random background, while our rendering is based on real images and the synthesized images are more photo-realistic. For FineNet training, we render images with fine-scale details, and train FineNet in a supervised manner, while~\cite{richardson2016learning} trains FineNet in an unsupervised manner. 2) For easy back propagation, ~\cite{richardson2016learning} adopts the first-order spherical harmonics (SH) to model lighting, while we use the second-order SH, which can reconstruct more accurate geometry details. 3) Our proposed loss function in CoarseNet better fits the goal and calculating the parameters in pixel level can achieve more stable and faster convergence. We did an experiment to compare our loss function $\mathcal{L}$ in Eq.~\eqref{eq:L} with the one used in~\cite{richardson2016learning}. Specifically, we used the two loss functions separately to train CoarseNet with 15000 iterations and batch size 100. Table~\ref{tab:loss_compare} shows the results of the test errors under different metrics on the test set (about 700 AFLW images). We can see that no matter which metric is used, either our defined metrics ($\mathcal{L}_{\textrm{pose}}$ and $\mathcal{L}_{\textrm{geo}}$), or the metrics employed in~\cite{richardson2016learning} (MSE for pose parameters and vertex distance for geometry parameters), our method always achieves lower testing errors than~\cite{richardson2016learning}, which demonstrates the effectiveness of the defined loss function for training.

\begin{table}
 \centering
 \caption{Comparisons of testing errors under different metrics.}  \label{tab:loss_compare}
\begin{tabular}{|c|c|c|}
\hline
Metrics     &~\cite{richardson2016learning}     &Our method \\ \hline
$\mathcal{L}_{\textrm{pose}}$ in Eq.~\eqref{eq:Lpose} &26.35  &7.69 \\ \hline
$\mathcal{L}_{\textrm{geo}}$ in Eq.~\eqref{eq:Lgeo} &5.53   &4.23 \\ \hline
MSE (pose parameters) &1.91   &0.56 \\ \hline
Mean vertex distance (geometry parameters) &5.18   &4.55 \\ \hline
\end{tabular}
\end{table}

\para{Quantitative results of single-image based dense face reconstruction}
For quantitative evaluation, we compare our method with the landmark-based method~\cite{zhu2015high} and the learning-based method~\cite{zhu2016face} on the Spring2004range subset of Face Recognition Grand Challenge dataset V2~\cite{phillips2005overview}. The Spring2004range has 2114 face images and their corresponding depth images. We use the face alignment method~\cite{kazemi2014one} to detect facial landmarks as the input of~\cite{zhu2015high}. For comparison, we project the reconstructed 3D face on the depth image, and use both Root Mean Square Error (RMSE) and Mean Absolute Error (MAE) metrics to measure the difference between the reconstructed depth and the ground truth depth on the valid pixels. We discard some images in which the projected face regions are very far away from the the real face regions for any of the three methods, which leads to a final 2100 images being chosen for the comparisons. The results are shown in Table~\ref{tab:quant}. It can be seen that our method outperforms the other two recent methods in both RMSE and MAE. The results of~\cite{zhu2015high} and~\cite{zhu2016face} are generated by directly running their released codes in public.

\begin{table}[!t]
 \centering
 \caption{Quantitative Comparison. Our method outperformes \cite{zhu2015high} and \cite{zhu2016face} in terms of RMSE and MAE.}  \label{tab:quant} \vspace{0.1in}
\begin{tabular}{|c|c|c|}
\hline
Method & RMSE [mm]  & MAE [mm] \\
\hline
\cite{zhu2015high}  & 5.946 & 4.420 \\
\hline
\cite{zhu2016face} & 5.367 & 3.923 \\
\hline
Ours  & 4.915 & 3.846 \\
\hline
\end{tabular}
\end{table}

Note that we are not able to perform a quantitative comparison with the state-of-the-art method~\cite{richardson2016learning}, since their code is not released. Their reported MAE value for the Spring2004range dataset is lower than what we obtain in Table~\ref{tab:quant}. We believe it is due to the masks they used in their MAE computation, which are unfortunately not available to us. Although we cannot give a quantitative comparison, the visual comparison shown in Fig.~\ref{fig:compare} clearly demonstrates the superior face reconstruction performance of our method.


\section{Conclusions}
We have presented a coarse-to-fine CNN framework for
real-time textured dense 3D face reconstruction and tracking from
monocular RGB video as well as from a single RGB image. The training data to our convolutional networks
are constructed by the optimization based inverse rendering
approach. Particularly, we construct the training data by varying the
pose and expression parameters, detail transfer as well as simulating
the video-type adjacent frame pairs. With the well constructed
large-scale training data, our framework recovers the detailed geometry,
albedo, lighting, pose and projection parameters in real-time. We believe that our well constructed datasets including 2D face images, 3D coarse face models, 3D fine-scale face models, and multi-view face images of the same person could be applied to many other face analysis problems like face pose estimation, face recognition and face normalization.

\edit{Our work has limitations. Particularly, like many recent 3D face reconstruction works~\cite{shi2014automatic,ichim2015dynamic,garrido2016reconstruction}, we assume Lambertian surface reflectance and smoothly varying illumination in our
inverse rendering procedure, which may lead to inaccurate fitting for face images with specular reflections or self-shadowing. It is worth to investigate more powerful formulation to handle general reflectance and illumination.}

\vspace{-2mm}
\section*{Acknowledgments}
We thank Thomas Vetter et al. and Kun Zhou et al. for allowing us to use their 3D face datasets. This work was supported by the National Key R\&D Program of China (No. 2016YFC0800501), the National Natural Science Foundation of China (No. 61672481).

{\small
\bibliographystyle{ieee}
\bibliography{egbib}
}

\begin{IEEEbiography}[{\includegraphics[width=1in]{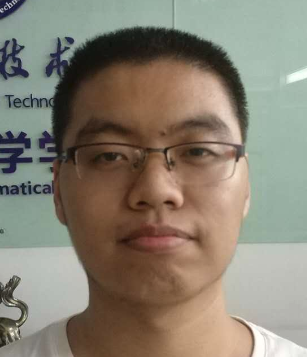}}]{Yudong Guo}
is a master student at the School of Mathematical Sciences, University of Science and Technology of China. He obtained his bachelor degree from the same University in 2015. His research interests include Computer Vision and Computer Graphics.
\end{IEEEbiography}

\vspace{-5mm}
\begin{IEEEbiography}[{\includegraphics[width=1in]{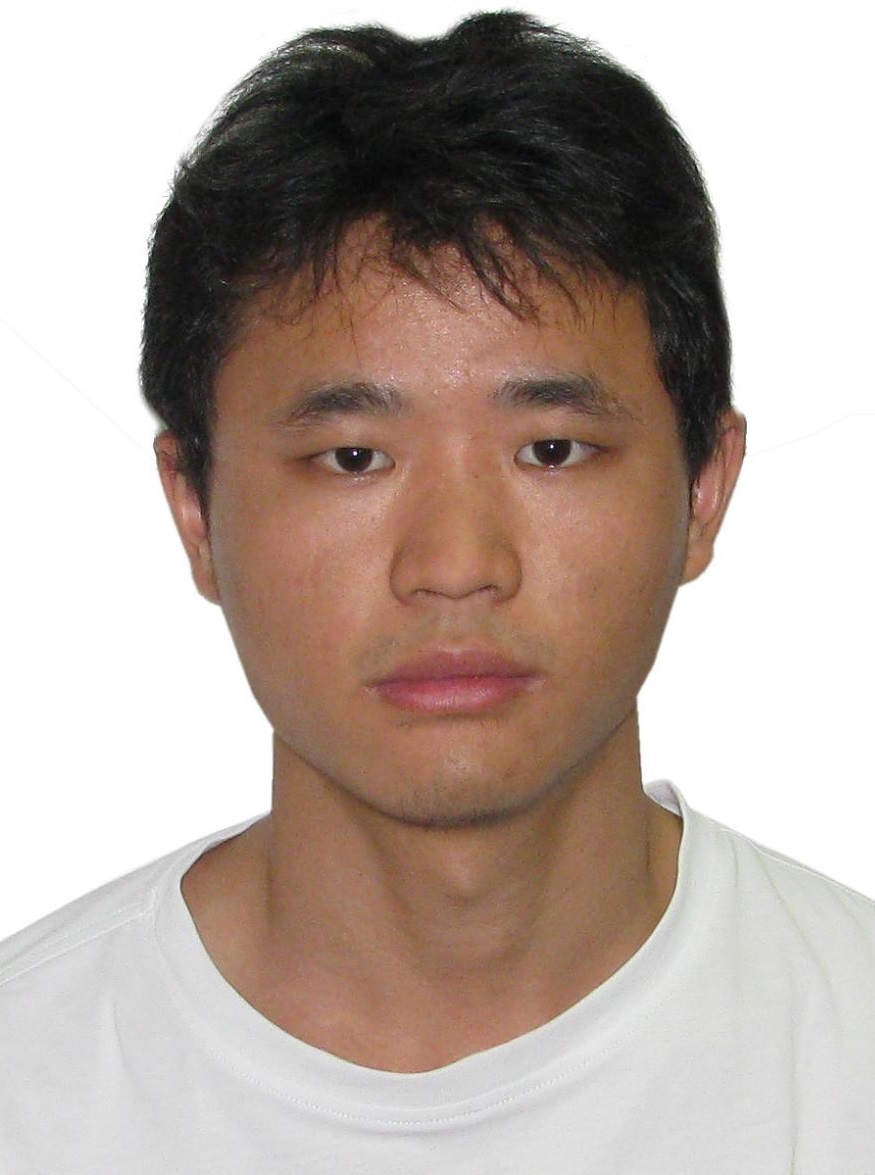}}]{Juyong Zhang}
is an associate professor in the School of Mathematical Sciences at University of Science and Technology of China. He received the BS degree from the University of Science and Technology of China in 2006, and the PhD degree from Nanyang Technological University, Singapore. His research interests include computer graphics, computer vision, and numerical optimization. He is an associate editor of The Visual Computer.
\end{IEEEbiography}

\vspace{-5mm}
\begin{IEEEbiography}[{\includegraphics[width=1in]{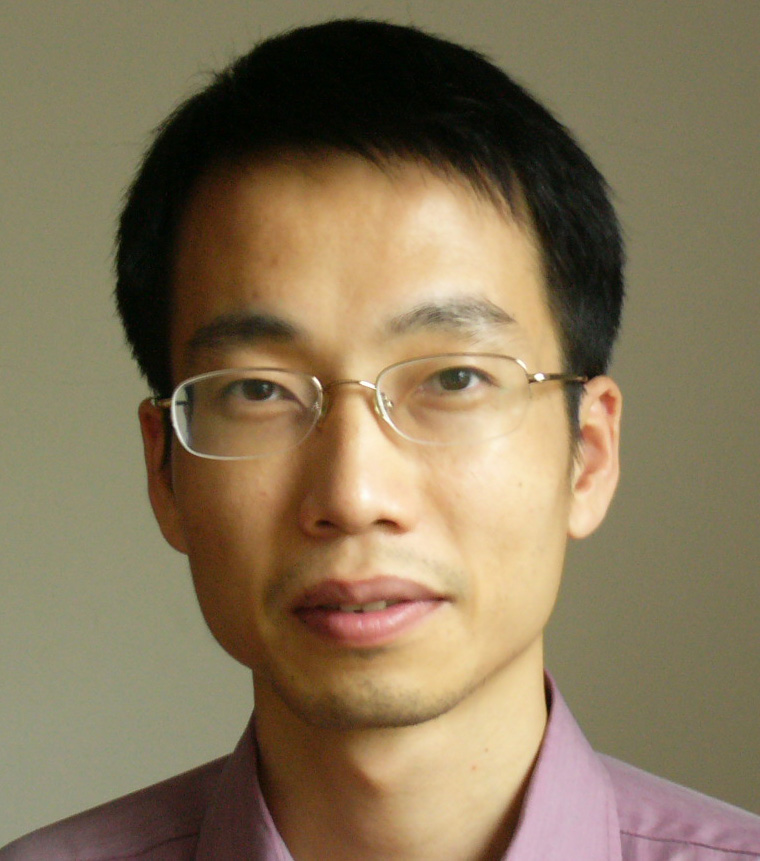}}]{Jianfei Cai}(S'98-M'02-SM'07)
received his PhD degree from the University of Missouri-Columbia. He is currently an Associate Professor and has served as the Head of Visual \& Interactive Computing Division and the Head of Computer Communication Division at the School of Computer Engineering, Nanyang Technological University, Singapore. His major research interests include multimedia, computer vision and visual computing. He has published over 200 technical papers in international journals and conferences. He is currently an AE for IEEE TMM, and has served as an AE for IEEE TIP and TCSVT.
\end{IEEEbiography}

\vspace{-5mm} \begin{IEEEbiography}[{\includegraphics[width=1in]{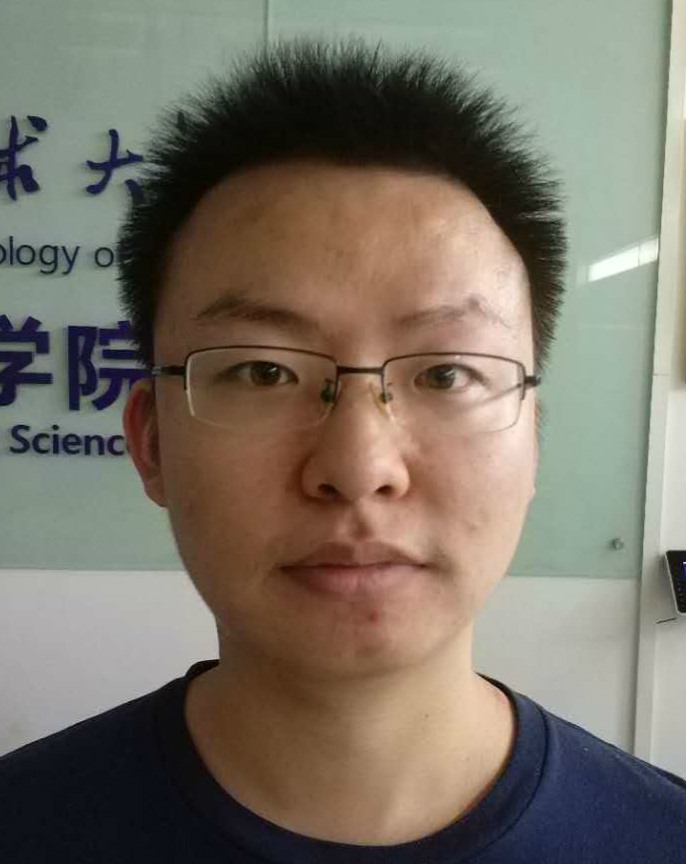}}]{Boyi Jiang}
received his bachelor degree in Mathematical Sciences from University of Science and Technology of China in 2015. He is now pursuing his Master degree in the same University. His research interests are Computer Vision and Digital Geometry Processing.
\end{IEEEbiography}

\vspace{-5mm}
\begin{IEEEbiography}[{\includegraphics[width=1in]{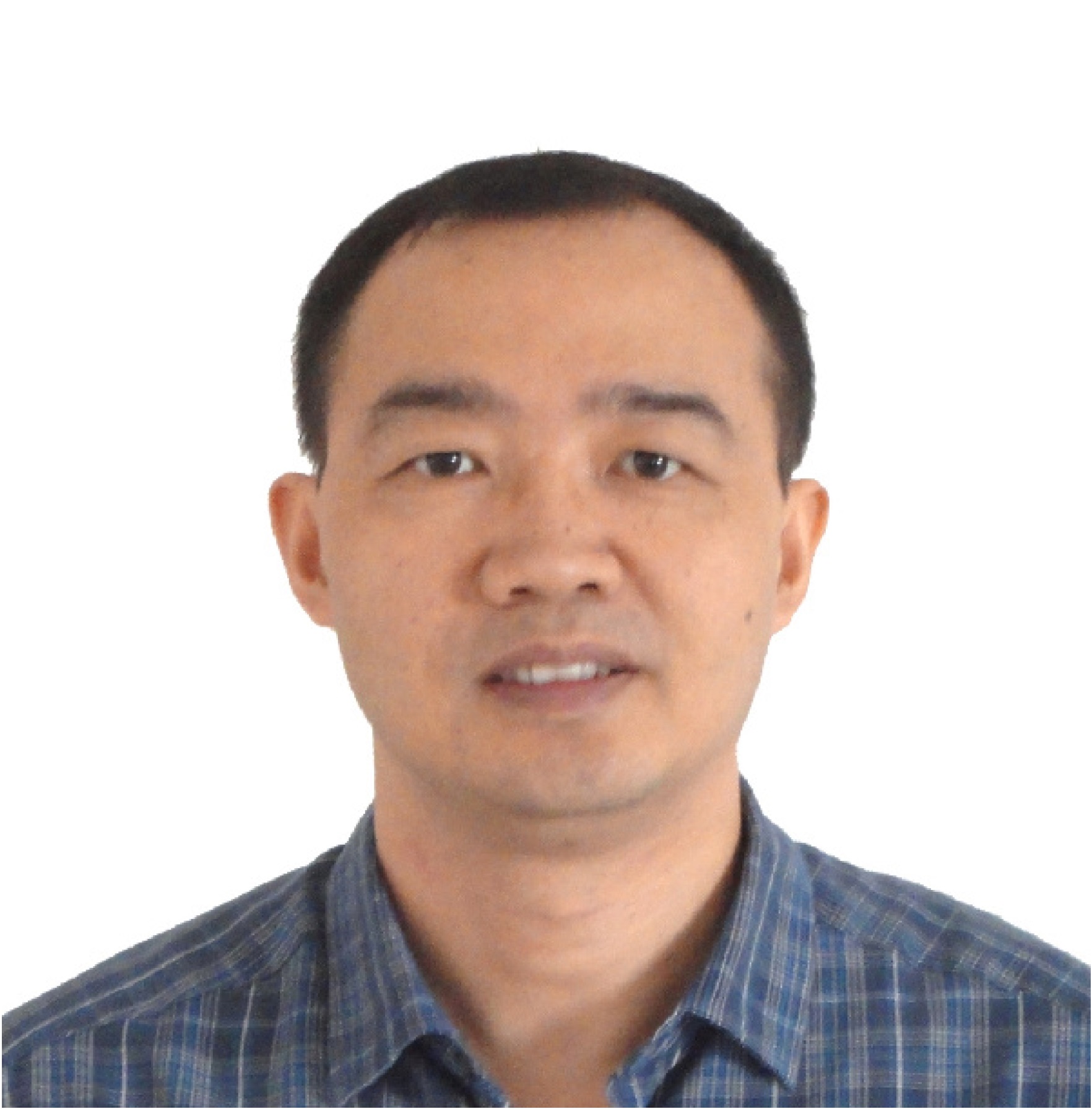}}]{Jianmin Zheng}
is an associate professor in the School of Computer Engineering at Nanyang Technological University, Singapore. He received the BS and PhD degrees from Zhejiang University, China. His recent research focuses on T-spline technologies, digital geometric processing, reality computing, 3D vision, interactive digital media and applications. He has published more than 150 technical papers in international conferences and journals. He was the conference co-chair of Geometric Modeling and Processing 2014 and has served on the program committee of several international conferences. He is an associate editor of The Visual Computer.
\end{IEEEbiography}

\onecolumn

\pagebreak
\begin{center}
\textbf{\large Supplementary Material}
\end{center}

This supplementary material shows the experimental result comparisons with the-state-of-the-art methods on 19 test images.

\begin{figure}[!b]
\includegraphics[width=1\textwidth]{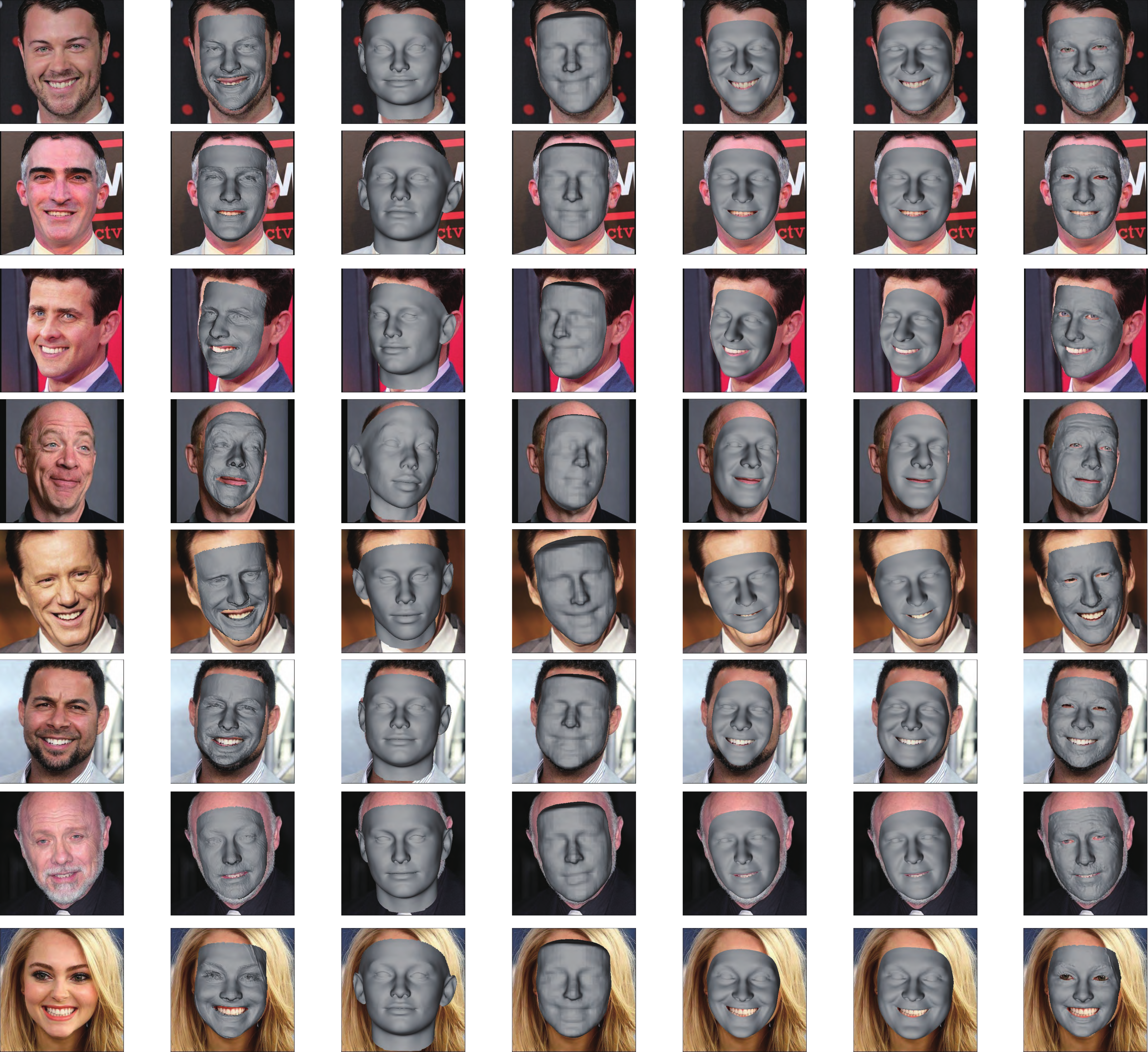}
Input \qquad \qquad \qquad [42] \qquad \qquad \qquad [3] \qquad \qquad \qquad [24] \qquad \qquad \qquad [46] \qquad \qquad \qquad [14] \quad \qquad \qquad Ours
\caption{Detailed comparisons with the state-of-art methods [42,3,24,46,14]. It can be seen that our results are more convincing in both the global geometry and the fine-scale details.}
\end{figure}

\begin{figure*}[!b]
\includegraphics[width=1\textwidth]{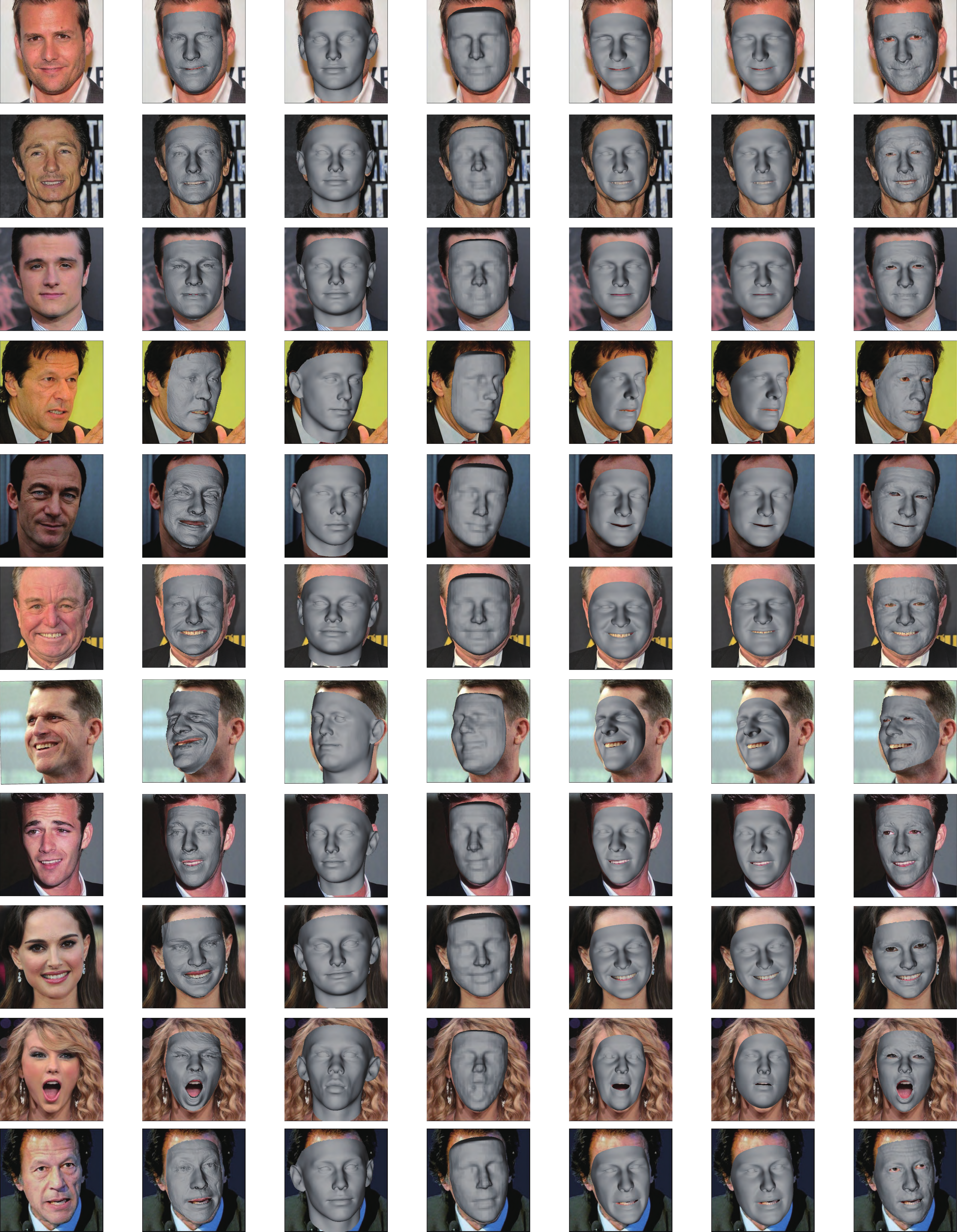}
Input \qquad \qquad \qquad [42] \qquad \qquad \qquad [3] \qquad \qquad \qquad [24] \qquad \qquad \qquad [46] \qquad \qquad \qquad [14] \quad \qquad \qquad Ours
\caption{Detailed comparisons with the state-of-art methods [42,3,24,46,14]. It can be seen that our results are more convincing in both the global geometry and the fine-scale details.}
\end{figure*}


\end{document}